%% file: main.tex
\documentclass[final,2p,times,authoryear]{elsarticle}

\makeatletter
\def\ps@pprintTitle{%
 \let\@oddhead\@empty
 \let\@evenhead\@empty
 \def\@oddfoot{\centerline{\thepage}}%
 \let\@evenfoot\@oddfoot}
\makeatother
\usepackage{amssymb}
\usepackage{booktabs}
\usepackage{subcaption}
\usepackage{graphicx}
\usepackage{multirow}
\usepackage{makecell}
\usepackage{array}
\usepackage{makecell}

\input{math_commands.tex}

\usepackage{hyperref}
\usepackage{xcolor}
\usepackage[super]{nth}
\usepackage{adjustbox}

\usepackage{lineno}

\journal{Remote Sensing of Environment}

\begin{document}

\begin{frontmatter}


\title{Crop mapping from image time series: deep learning with multi-scale label hierarchies}

\author[a]{Mehmet Ozgur Turkoglu} 
\author[a]{Stefano D'Aronco} 
\author[b]{Gregor Perich}
\author[b,c]{Frank Liebisch}
\author[d]{Constantin Streit}
\author[a]{Konrad Schindler}
\author[a,e]{Jan Dirk Wegner}
\address[a]{EcoVision Lab,\ Photogrammetry and Remote Sensing,\ ETH Zurich, Switzerland\\}
\address[b]{Crop Science,\ ETH Zurich, Switzerland\\}
\address[c]{Agroecology and Environment, Agroscope, Switzerland\\}
\address[d]{Federal Office for Agriculture, Switzerland\\}
\address[e]{Institute for Computational Science, University of Zurich, Switzerland\\}

\begin{abstract}
The aim of this paper is to map agricultural crops by classifying satellite image time series.
Domain experts in agriculture work with crop type labels that are organised in a hierarchical tree structure, where coarse classes (like \emph{orchards}) are subdivided into finer ones (like \emph{apples}, \emph{pears}, \emph{vines}, etc.).
We develop a crop classification method that exploits this expert knowledge and significantly improves the mapping of rare crop types. The three-level label hierarchy is encoded in a convolutional, recurrent neural network (convRNN), such that for each pixel the model predicts three labels at different level of granularity. This end-to-end trainable, hierarchical network architecture allows the model to learn joint feature representations of rare classes (e.g., \emph{apples}, \emph{pears}) at a coarser level (e.g., \emph{orchard}), thereby boosting classification performance at the fine-grained level. Additionally, labelling at different granularity also makes it possible to adjust the output according to the classification scores; as coarser labels with high confidence are sometimes more useful for agricultural practice than fine-grained but very uncertain labels.
We validate the proposed method on a new, large dataset that we make public. \emph{ZueriCrop} covers an area of 50 km $\times$ 48 km in the Swiss cantons of Zurich and Thurgau with a total of 116'000 individual fields spanning 48 crop classes, and 28,000 (multi-temporal) image patches from Sentinel-2. 
%
We compare our proposed hierarchical convRNN model with several baselines, including methods designed for imbalanced class distributions. The hierarchical approach performs superior by at least 9.9 percentage points in F1-score. 

\end{abstract}


\end{frontmatter}

\section{Introduction}
\label{Introduction}
\input{00_intro}

\section{Related Work}
\label{Related Work}
\input{01_related_work}

\section{Method}
\label{Method}
\input{02_method}

\section{Dataset}
\label{Dataset}
\input{03_dataset}

\section{Experiments}
\label{Experiments}
\input{04_experiments}
\section{Results}
\label{Results}
\input{05_result}

\section{Discussion}
\label{Discussion}
\input{06_discussion}

\section{Conclusion}
\label{Conclusion}

\input{07_conclusion}

\section*{Acknowledgments}
We thank the Swiss Federal Office for Agriculture (FOAG) for partially funding this Research project through the DeepField Project.

\vfill


\clearpage
  \bibliographystyle{elsarticle-harv} 
  \bibliography{bib}
\clearpage

\appendix
\input{08_appendix}

\end{document}

%% file: math_commands.tex
\usepackage{amsmath,amsfonts,bm}

\def\eqref#1{equation~\ref{#1}}

\def\1{\bm{1}}

\DeclareMathAlphabet{\mathsfit}{\encodingdefault}{\sfdefault}{m}{sl}
\SetMathAlphabet{\mathsfit}{bold}{\encodingdefault}{\sfdefault}{bx}{n}
\newcommand{\tens}[1]{\bm{\mathsfit{#1}}}

\def\tB{{\tens{B}}}

\def\tH{{\tens{H}}}

\def\tK{{\tens{K}}}

\def\tW{{\tens{W}}}
\def\tX{{\tens{X}}}
\def\tY{{\tens{Y}}}
\def\tZ{{\tens{Z}}}



%% file: 00_intro.tex
Monitoring agricultural land use is of high importance for food production, biodiversity, and forestry \citep{Gomez2016}. An increasing world population, climate change, and changes in food consumption habits put yet uncultivated areas under pressure, while leading to intensification in existing agricultural areas \citep{Laurance2014}. 
Cropland expansion and intensive use of agricultural areas are often connected with negative ecological impacts like deforestation and biodiversity loss, but also degradation of ecosystem services like ground and surface water quality \citep{Herzog2008, Dise2011}. Therefore, dense, accurate monitoring of agricultural lands plays an essential role for their optimal and sustainable management.
Knowledge of crop areas and certain land uses is of importance for many political programs that aim to reduce and alleviate the environmental impacts of intensive agriculture, too \citep{Gomez2016}. Policy driven incentives, for instance, foster a particular share of a farm area to remain extensively used grassland to promote biodiversity, or give subsidies to promote a certain crop mix in the rotation \citep{Finger2012}. 
Collecting information is traditionally based on farmer self-reporting and spot checking by the authorities in the field, which is laborious, costly, and prone to errors. 

Modern machine learning methods in combination with publicly available satellite imagery provide new possibilities for more accurate spatially dense monitoring of agricultural sites at high temporal resolution and low cost. One particularly promising recent sensor is Sentinel-2, due to its low ground sampling distance (10 m) at a revisit rate of 3-5 days.
In general, the spectral signal of the vegetation as captured by the satellite has specific characteristics as a function of \emph{(i)} soil structure and composition (e.g., soil brightness, soil water content, soil type, etc.), \emph{(ii)} vegetation structure (e.g., canopy cover, Leaf Area Index (LAI), plant height, leaf angle, etc.) and \emph{(iii)} leaf biochemistry (e.g., chlorophyll, water content, nitrogen content, etc.)~\citep{Thenkabail2013}. 
Not only each plant species has its own spectral signature, but spectral characteristics are also highly dependent on the phenological stage of the plant~\citep{walter2015pheno, anderegg2020spectral}. Instead of merely analysing images at a single point in time, time-series (sequences) analysis of satellite images thus provides significant additional evidence about crop species.

Supervised machine learning -- recently in particular deep learning \citep{marc_lstm, bigru, fcn, zhong2019deep, tempCONV, breizhcrops, garnot2019time, pixel_set} has shown good performance as a tool for multi-temporal vegetation mapping, on different datasets \citep{marc_lstm, bigru, fcn, zhong2019deep, tempCONV, breizhcrops, pixel_set}. 
However, these existing datasets usually contain only a small number of relatively well-balanced crop classes (e.g., 9 classes in~\cite{breizhcrops}, 10 classes in~\cite{tempCONV}, 13 in~\cite{zhong2019deep}).
In practice, large-scale datasets that cover all existing plots in some geographic region have highly skewed and long tail class distributions with a large number of different classes.

When dealing with imbalanced data, a large number of (rare) classes comes with very few labels, which makes training any data-driven method challenging. A viable way to alleviate this problem consists in using hierarchical classification strategies~\citep{srivastava2013discriminative, b-cnn,hierarchical_network, cnn_rnn_combined, tree_cnn}.
%
%
Although many classes are rare, they may belong to the same super-class at a coarser level and share common features, e.g., leopards, tigers, lions, and cheetahs all share the visual properties of cats. This observation can be used to regularize the training of a model and improve the generalization error, especially for the rare classes. Imposing prior knowledge about the class structure adds (soft) constraints to the model and encourages it to pool shared information from related classes. Such "coarse-granularity features" are easier to learn due to the larger training set, while the fine-grained classification can focus on discriminating fewer sub-classes, thus using the rare training examples more efficiently. Crop classification is a task for which we can define such a label hierarchy. For example, apple orchards, pear orchards and chestnut orchards (which rarely appear individually) all belong to the same \emph{orchards} subclass and share many visual features. We point out another advantage of a hierarchical scheme: one can use the class scores to determine how reliable predictions are at different hierarchy levels. In case output scores at the most fine-grained level (e.g., apple orchard, pear orchard, chestnut orchard) are low, we can always use coarser outputs (orchard), which typically receive significantly higher scores because they aggregate evidence across more data. 
For many applications (e.g., summary statistics, calculation of subsidies) that coarser granularity of annotation is good enough. Moreover, only passing the fine-grained decisions between few, rare classes to human experts greatly reduces their manual cleaning and relabeling workload.

In this work, we propose a deep learning network architecture for crop mapping that is \emph{hierarchical}, to exploit a tree-structured label hierarchy built by domain experts; \emph{convolutional} to encode image data; and \emph{recursive} to represent time series.
The proposed architecture has multiple levels of representation that consist of stacked, convolutional recurrent neural networks. The different network levels predict successively finer label resolution in the hierarchical tree. 
%
%
Moreover, we add a label refinement module, which takes the predictions as input and refines them with a Convolutional Neural Network (CNN) that exploits the correlations between labels across the hierarchy.

In order to test our model, we introduce a new dataset called \emph{ZueriCrop}. This dataset is based on farm census data from the Swiss Federal Office for Agriculture (FOAG) and consists of annotated field polygons from the Cantons of Zurich and Thurgau from the year 2019. This dataset contains 48 different classes with a realistic, highly imbalanced class distribution. The dataset comes with a label hierarchical tree, built using expert knowledge, that can be leveraged during training.
Our experiments show that the proposed model outperforms the state-of-the-art methods on our \emph{ZueriCrop} dataset; in addition, it is more effective than widely used techniques for coping with imbalanced data distribution, such as data augmentation or class-balanced loss functions.
%
To summarize, our contributions are:
\begin{itemize}
  \item We propose a new, multi-temporal crop classification method that encodes a domain-specific label hierarchy directly inside an end-to-end trainable model architecture. It outputs labels at multiple granularity levels for each pixel and significantly improves classification accuracy. 
  \item We provide a new, publicly available crop classification dataset \emph{ZueriCrop}, equipped with a tree-structured label hierarchy. \emph{ZueriCrop} covers a 50 km $\times$ 48 km area in the Swiss cantons of Zurich and Thurgau. It contains 28,000 Sentinel-2 image patches of size 24 pixels $\times$ 24 pixels, each observed 71 times over a period of 52 weeks; 48 agricultural land cover classes; and 116,000 individual agricultural fields. 
\end{itemize}

%% file: 01_related_work.tex
\textit{Crop Classification with multi-temporal satellite data} has been widely studied in remote sensing. Traditional machine learning approaches with handcrafted features~\citep{inglada2015assessment, wardlow2008large, vuolo2018much} predominantly rely on vegetation indices like the Normalized Difference Vegetation Index (NDVI)~\citep{foerster2012crop, ustuner2014crop, pena2011object, conrad2010per}. 
%
Different strategies have been explored to better model the temporal evolution as further evidence for classification, such as temporal windows~\citep{conrad2014derivation}, hidden Markov models and dynamic time warping~\citep{siachalou2015hidden,belgiu2018sentinel}, and conditional random fields~\citep{bailly2018crop}. 
For instance,~\citet{wardlow2008large} extract features, which are time series of NDVI, from MODIS data collected over the growing season of crops; and they perform classification with a decision tree. Similarly~\citet{conrad2014derivation} investigate the optimum number of acquisition dates and most suitable temporal windows for the discrimination of crops from RapidEye satellite time-series data.
%
%
These traditional methods have in common that their performances are constrained by the limited discriminativeness and robustness of the hand-crafted features, as well as by the limited expressive power of conventional classifiers.

%
More recently deep learning methods have shown their ability to effectively solve many pattern recognition task. Their main advantage is twofold: \emph{(i)} they no longer rely on hand-engineered features to encode spectral, spatial, or temporal patterns; \emph{(ii)} their large capacity makes them able to learn very complex, highly non-linear relationships, if given sufficient labeled training data and computational resources. 
\citet{marc_lstm} use a recurrent neural network with Long Short-Term Memory (LSTM) to encode temporal dependencies in the data, while \citet{bigru} improve the result on the same dataset by encoding both, temporal \emph{and} spatial dependencies via convolutional LSTM and Gated Recurrent Units (GRUs). In \citep{fcn, garnot2019time}, satellite images are first processed individually with a CNN to obtain per-image features; then temporal dependencies between these features are modeled with a separate Recurrent Neural Networks (RNNs). Further options are temporal CNNs that combine features also across time with convolutions~\citep{tempCONV}, or models that use the attention principle~\citep{transformer} to aggregate information across time~\citep{breizhcrops, russwurm2020self}. \citet{pixel_set} combine pixel-set encoder and transformer~\citep{transformer} and show improved performance over RNN-based approaches.
Finally, in our previous work~\citep{star} we build a deep RNN with a new cell structure termed STAR that trains better than LSTM- and GRU-type models while being more parameter-efficient. This makes it possible to train deeper models, which translates to improved performance across a range of sequence modelling tasks, including crop classification.

\textit{Handling imbalanced datasets} is generally an issue in supervised classification. Modern deep learning methods are data-hungry and prone to overfit. In order to generalize well on the test data, a large amount of labeled training data is usually required. 
In practice, however, some classes occur more often than others (e.g., animal species, crop types) or some labels are simply easier to collect. This  leads to long-tailed class distributions being the norm, rather than the exception, for large, real-world datasets~\citep{sun_dataset}. 
Under standard training regimes, machine learning models tend to ignore rare, under-represented classes and focus on the dominant classes to maximize cumulative performance across the entire dataset~\citep{wang2017learning,ren2018learning,dong2018imbalanced}. While those shortcomings are easily overlooked when evaluating with global performance metrics like overall accuracy, they become obvious with class-balanced metrics like average class precision or F1-score. In fact, for many practical applications a class-balanced evaluation is essential, as rare classes have the same (or even higher) importance as frequent ones. This is also true for our agricultural mapping problem where crops that have high financial or ecological value (for instance orchards and vegetables) are rare compared to pastoral grasslands or wheat fields.

Two major strategies have been explored to counter class imbalance: \emph{(i)} algorithm-level approaches and \emph{(ii)} data-level approaches. A typical algorithm-level approach is cost-sensitive learning where the loss function is re-weighted by a factor inversely proportional to the class frequencies~\citep{ling2008cost,huang2016learning,khan2017cost,khan2019striking}, where training samples of rare classes receive higher weight. An inherent consequence of resampling or reweighting training samples according to rarity is a model bias towards the rare classes.
Data-level approaches try to balance the dataset either by oversampling minority classes~\citep{chawla2002smote,douzas2018effective,cb_beta} or by under-sampling the majority classes~\citep{he2009learning}. Undersampling dominant classes runs the risk to miss large parts of the data distribution, thus hurting model performance. On the other hand, oversampling rare classes reaches its limit if the number of available samples in a class is too low.
%
We propose to leverage the hierarchical structure of the labels to counter class imbalance and to improve performance for rare classes.
%
%

\textit{Hierarchical classification in remote sensing} has been investigated before. \citet{melgani2004classification} develop hierarchical tree-based classification strategies using binary classifiers like support vector machines (SVM) for hyper-spectral remote sensing data. 
\citet{chen2009hierarchical} propose a rule-based method that hierarchically classifies land cover and land use from LIDAR and WorldView-2 data. Similarly, \citet{wu2016hierarchical} apply a rule-based classification of LIDAR data for \emph{building} classification followed by a classification of \emph{road}, \emph{vegetation}, and \emph{bare soil} with SVMs by additionally incorporating WorldView-2. 
Another rule-based, hierarchical method is proposed in \citep{heupel2018progressive} for crop classification from four different satellite sensors: Landsat-7 and Landsat-8, Sentinel-2A and Rapid-Eye. The first-level classifier decides whether it is \emph{winter crop} or \emph{summer crop} and the second-level classifiers predict into eight fine-grained classes, e.g., \emph{potato, corn}. Their method exhaustively relies on hand-crafted features and expert knowledge. 
\citet{jiao2019hierarchical} apply rule-based decision-making to classify land covers of coastal wetlands into four coarse classes 
which are subdivided into more fine-grained classes with SVMs. 
%
\citet{goel2018hierarchical} study hierarchical metric learning for classification of remote sensing data. They use iterative max-margin clustering to organize the classes in a hierarchical fashion and subsequently learn different distance metric transformations for the classes present at the non-leaf nodes of the tree.
Another idea is using individual Random Forests at different hierarchy levels for land cover and land use classification \citep{ sulla2011hierarchical,sulla2019hierarchical} from MODIS data. More recently, \citet{demirkan2020hierarchical} investigated the benefit of hierarchical classification with SVMs and Random Forests for land cover and land use classification from Sentinel-2 data.
Although hierarchical classification in remote sensing has been studied before, all methods use traditional, hand-crafted features and decision trees designed for specific scenes and datasets by domain experts. Complex workflows that employ multiple independent classifiers at different stages are costly to compute, need much manual tuning for each new dataset, and erroneous decisions at early stages can hardly be compensated for later on in the pipeline. Additionally, none of the existing workflows dealt with a hierarchical approach for a realistic, large-scale, imbalanced dataset with a large number of classes.


\textit{Hierarchical classification} has already been studied in deep learning literature. \citet{srivastava2013discriminative} introduce a tree-like hierarchy in CNNs for image classification. Their method learns to organize the classes into a tree hierarchy that imposes a prior over the classifier’s parameters, which improves performance for minority classes.
\citet{yan2015hd} embed deep CNNs into a two-level hierarchy. Easily distinguishable classes are separated with a coarser classifier, while another classifier separates the other, more difficult cases at a more fine-grained level.
\citet{xiao2014error, tree_cnn} study hierarchical networks composed of deep CNNs in the context of incremental learning.
\citet{complement_obj} propose a training strategy that leverages the information from a label hierarchy. It maximizes the probability of the ground truth class, and at the same time, neutralizes the probabilities of the other classes in a hierarchical fashion, making the model take advantage of the label hierarchy explicitly.
Another interesting recent work, in the field of (hierarchical) text classification, is \citep{hilap}, where the hierarchy is explored at both training and inference time with a Markov decision process and deep reinforcement learning.
\citet{b-cnn} and \citet{hierarchical_network} propose multi-stage deep CNNs. Like ours, their models have multiple outputs of different granularity as well as multiple objectives. \citet{cnn_rnn_combined} further improve on that idea by combining a CNN that extracts a hierarchical image representation with an RNN to capture the hierarchical tree of labels. 

To the best of our knowledge, our method is the first that explicitly encodes the inherent (and for domain experts well-known) label hierarchy for crop classification in satellite image sequences in the deep learning setting. Our method differs from all existing literature in that it is based on an integrated, convolutional and recurrent model (convRNN) that is able to capture all relevant spatio-temporal correlations.
%
%
To demonstrate these features, and to enable further work in this direction, we also provide a new dataset which, compared to existing ones~\citep{fcn, breizhcrops, marc_lstm}, has many more classes and a realistic, much less balanced class distribution.

%% file: 02_method.tex

\begin{figure}[ht]
    \centering
        \includegraphics[width=0.88\columnwidth]{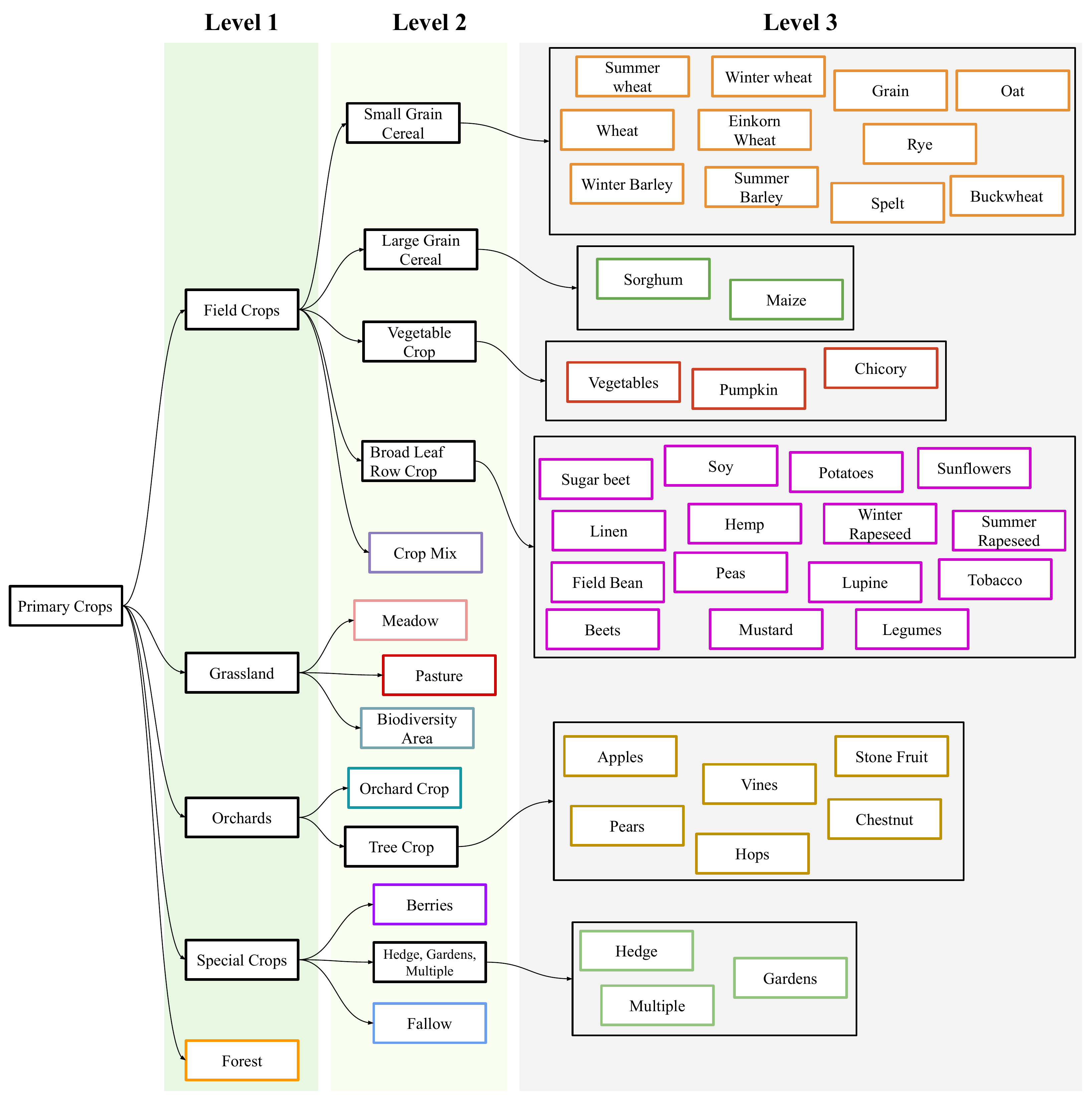}
    \caption{The hierarchy of all crop classes of the \emph{ZueriCrop} dataset. Black box indicates intermediate label levels while color boxes indicates the finest granularity.
    }
    \label{fig:label_hier}
\end{figure}

Formally, our objective is to predict a crop type map $\tY \in  {\rm I\!R}^{H\times W\times C}$ from a sequence of input images $\{\tX_1,\tX_2,..,\tX_T\} \in  {\rm I\!R}^{H\times W\times B}$. $H$ and $W$ are height and width of the input images, respectively, $B$ is the number of input bands, $T$ is number of time stamps in the input sequence, and $C$ is the number of crop types. See Fig.~\ref{fig:data_example}. 
%
We assume that multiple labels are assigned to a single pixel. Each of those labels belongs to a different level in a hierarchical structure that encodes agricultural crop types at a different granularity, from coarse to fine. This label hierarchy is created by human experts, the levels have an intrinsic semantic meaning. Fig.~\ref{fig:label_hier} shows our label tree for all classes used in our dataset. 

Labels are denoted as $\tY^n$, where $n\in{1,...,N}$ represents the level inside the label hierarchy. Even though there are multiple labels for each crop type, the ultimate goal is to predict (as much as possible) the finest granularity $\tY^N$, corresponding to detailed species labels. In our dataset, we distinguish three hierarchy levels, with 1 the coarsest and 3 the finest. We note that although, for clarity, we stick to the case  $N=3$ for the rest of the paper, our method is generic and be used for other values of $N$.
%
%

%
\subsection{Convolutional Recurrent Neural Networks}
Convolutional recurrent neural networks (convRNN) are the convolutional version of RNNs, designed to represent spatio-temporal data. ConvRNNs have been used for different spatio-temporal modeling tasks like weather forecasting \citep{convlstm}, video action recognition \citep{videolstm}, video forecasting \citep{tt-convlstm}, and the prediction of heat diffusion \cite{phys_incorporated}.
They differ from standard RNNs in that the matrix multiplications are replaced with the convolution operator. In general it is straightforward to convert any recurrent cell with its convolutional version, see for instance~\citep{convlstm,convGRU,star}. 
Constructing a network with multiple layers helps learning more discriminative evidence via a richer set of features. However, training networks with many layers is hard if using widely known LSTM and GRU cells as basic recurrent units as shown in~\citep{star}. A computationally more efficient cell type with less parameters that allows training deeper models is convSTAR~\citep{star}. We thus construct a network using convSTAR units and demonstrate its superior performance over versions using GRU and LSTM units in the experiments section (Section~\ref{Results}), Table~\ref{table:ablation_RNN}). In the following, we will briefly recap the design of a convSTAR cell before describing the construction of our hierarchical approach.


More formally, the convSTAR cell at convolution layer $l$ and at time $t$, takes as input the hidden state tensor $\tH_{t}^{l-1}$ of the previous layer $(l-1)$, where the input $\tH_{t}^0$ corresponds to the channels of the multi-spectral input image $\tX_{t}$.
That tensor is first non-linearly transformed and then linearly combined with the previous hidden state $\tH_{t-1}^l$, with the weights of the linear combination modulated by a gating variable $\tK^{l}_{t}$ that 
depends on both inputs and controls the information flow. Formally, the computation is given by
\begin{align}
&\tK_{t}^{l} = \sigma( \tW_x*\tH_t^{l-1} + \tW_h*\tH_{t-1}^{l} + \tB_K )\\
&\tZ_t^{l} =  \tanh(\tW_z*\tH_t^{l-1} + \tB_z)\\
&\tH_t^{l} = \tanh(\tH_{t-1}^{l} + \tK_{t}^{l}\circ\big(\tZ_t^{l}-\tH_{t-1}^{l})\big)
\end{align}
where $\sigma$ is the sigmoid non-linearity, $*$ denotes convolution, $\circ$ is the Hadamard (element-wise) product, and $\tW$ and $\tB$ are the trainable weight and bias tensors, $l$ and $t$ denote the cell layer and time step of the sequence, respectively. The hidden state $\tH^{L}_{T}$ of the deepest ($L$-th) layer serves as latent encoding of the input sequence up to time $T$ and can be used for classification, regression or forecasting, by passing it through an appropriate decoder. 

\subsection{Hierarchical Convolutional Recurrent Network}
\begin{figure}[t]
    \centering
        \includegraphics[width=1.\columnwidth]{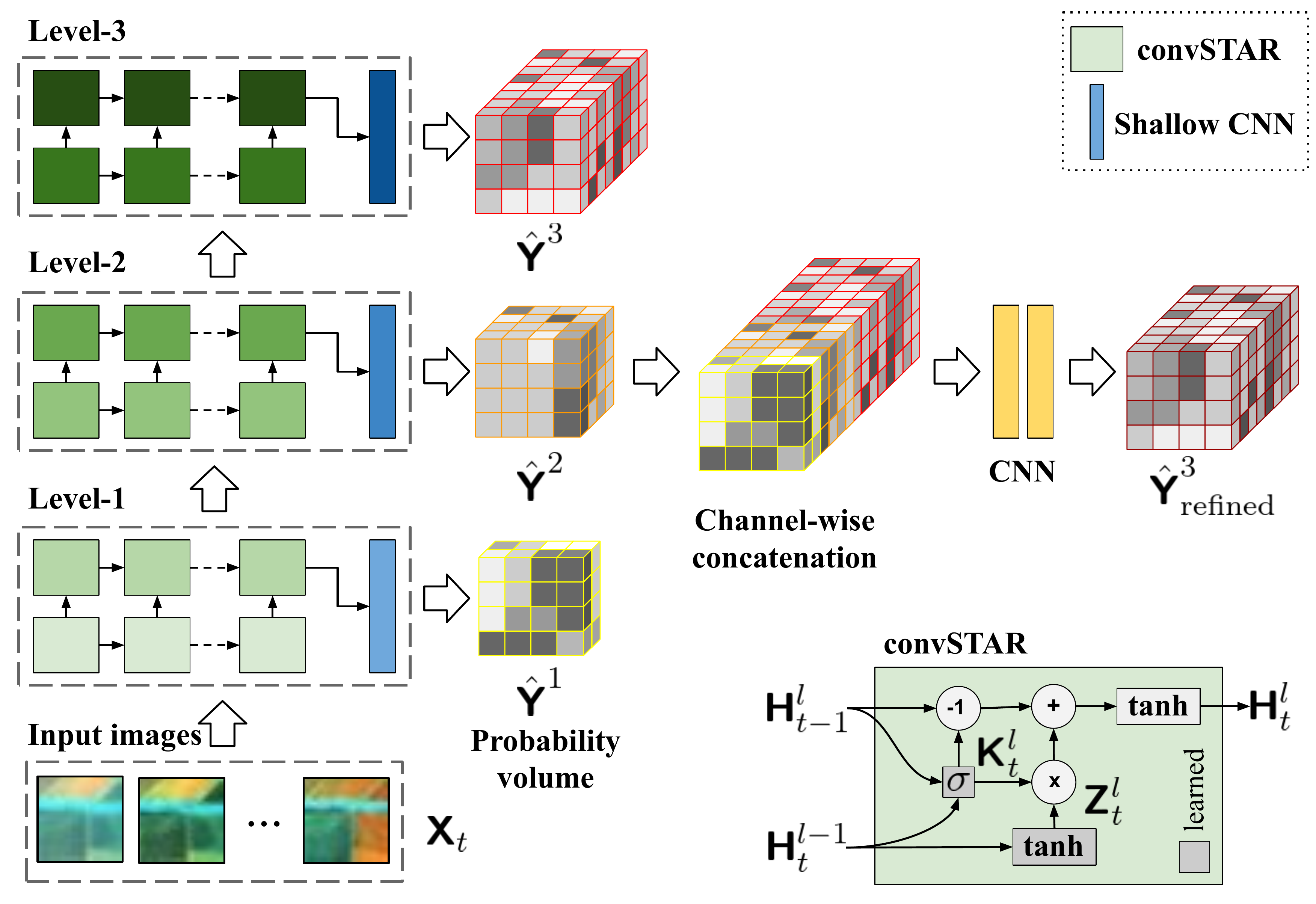}
    \caption{Our proposed hierarchical, multi-stage, convolutional STAR network (ms-convSTAR).}
    \label{fig:model}
\end{figure}

In order to construct a hierarchical representation similar to multi-stage CNN classification networks, but for image sequences instead of single images~\citep{hierarchical_network, b-cnn}, we propose a network that has $N$ stages. Each stage is made of a $2$-layer convSTAR architecture (Fig.~\ref{fig:model}). The output tensors (hidden states) at stage $n$ are fed as inputs to the next stage $(n+1)$. The final hidden state $(\tH_T^2)^n$ of each hierarchical level is fed to a conventional CNN classifier to obtain label scores $\hat{\tY}^n$.

All convolutional kernels have size $3\times3$. Each convSTAR layer has $64$ filters, a shallow 1-layer convolutional neural network (CNN) is used to convert the final hidden state to a patch of labels.
The network is trained with the cross-entropy ($CE$) loss, leading to the objective function
\begin{equation}
    L = \sum_{n=1}^{N} \lambda_n CE(\tY^n,\hat{\tY}^n) = - \sum_{n=1}^{N} \sum_{c=1}^{C} \lambda_n \tY^n_c \log(\hat{\tY}^n_c)
    \label{eq:loss}
\end{equation}
where $\lambda_n$ represents hyper-parameters that determine the relative influence of different hierarchy levels on the tree, with $\sum\lambda_n=1$.
By taking into account the loss at all levels of granularity, the network imposes the label hierarchy as a (soft) prior to guide the feature encoding. For example, the features of \emph{apple orchards} and \emph{pear orchards} should support an assignment to the coarser \emph{orchard} label, too.
%
\begin{figure}[t]
    \centering
        \includegraphics[width=1.\columnwidth]{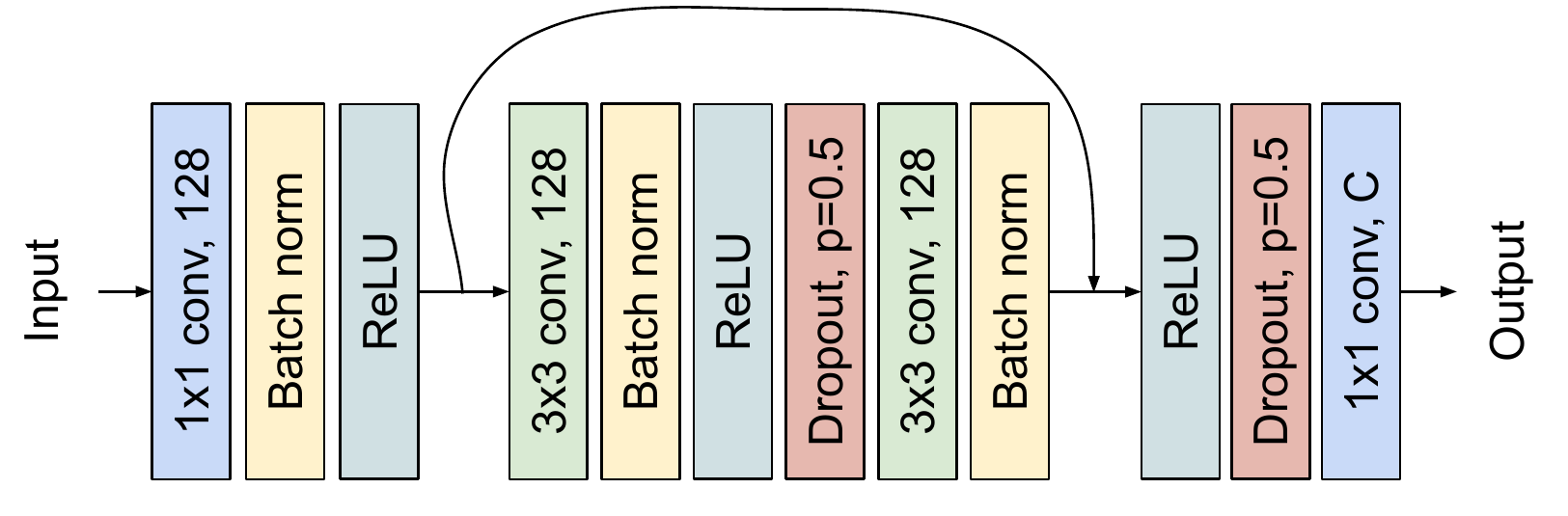}
    \caption{Label refinement CNN architecture.}
    \label{fig:cnn}
\end{figure}

\subsection{Label Refinement}
The model described so far embeds the label hierarchy in the network. However, by itself this does not guarantee consistency between the predictions at different stages. As an extreme example, the network could learn to simply ignore the input from the coarser hierarchy level and build independent classifiers for different levels of granularity. As a consequence, labels predicted at test time could possibly violate the parent-child relations of the hierarchy. A pixel could receive labels \emph{sunflower} and at the same time \emph{orchard}, for example.
To imprint the preference for coherent labels across the hierarchy levels, we add a \emph{label refinement network}, which is a CNN after the multi-level ms-convSTAR network. The refinement stage takes all the predictions $\{ \hat{\tY}^1\hdots \hat{\tY}^N\}$ from the individual network stages and produce a final prediction for the finest label granularity, $\hat{\tY}^N_{\text{refined}}$, while modelling their interactions. More specifically, 3-dimensional probability volumes from three stages are concatenated in the channel dimension and fed to the CNN. See Fig.~\ref{fig:model}. Formally, it is defined as 
\begin{equation}
    \hat{\tY}^N_{\text{refined}} = \hat{\tY}^N + F(\text{concat}[\hat{\tY}^1, \hat{\tY}^2,...,\hat{\tY}^N])
\end{equation}
where $F$ represents a CNN, see Fig.~\ref{fig:cnn}. 
The final loss function of the network is the weighted sum of the ms-convSTAR losses and the refinement loss, with a hyper-parameter $\gamma$ for the relative influence of the refinement loss:
\begin{equation}
    L = \sum_{n=1}^{N} \lambda_n CE(\tY^n,\hat{\tY}^n)  + \gamma CE(\tY^N,\hat{\tY}^N_{\text{refined}})
\end{equation}
where $\hat{\tY}^N_{\text{refined}}$ represent the output of the label refinement module and has exactly the same dimension as $\hat{\tY}^N$, see Fig.~\ref{fig:model}. 
Note that, empirically, we observed that refining coarser predictions does not affect the final model performance significantly. For experimental evaluation we thus run the refinement module exclusively for the finest label granularity. 

%
%
%

\subsection{Implementation details}

We have implemented our network architecture in PyTorch. Input image sizes are $H=W=24$, $B=4$ and sequence length is $T=71$. For all experiments, we use Adam~\citep{adam} as optimiser with batch size $4$ and run the training for $30$ epochs. The learning rate is set to $0.001$ at the start of the training and divided by $10$ every $10$ epochs. The model is regularised with weight decay of $0.0001$. Gradient magnitudes are clipped to $5$ to prevent exploding gradients. The hyper-parameters of the loss function~(\ref{eq:loss}) are empirically determined and set to $\lambda_1 = 0.1$, $\lambda_2 = 0.3$, and $\lambda_3 = \gamma = 0.6$. 
Input image patches are flipped randomly with 66\% chance during training for data augmentation.
All source code, trained models and the dataset are available online at \url{https://github.com/0zgur0/ms-convSTAR}.

%% file: 03_dataset.tex
\begin{figure}
    \centering
        \includegraphics[width=1.\columnwidth]{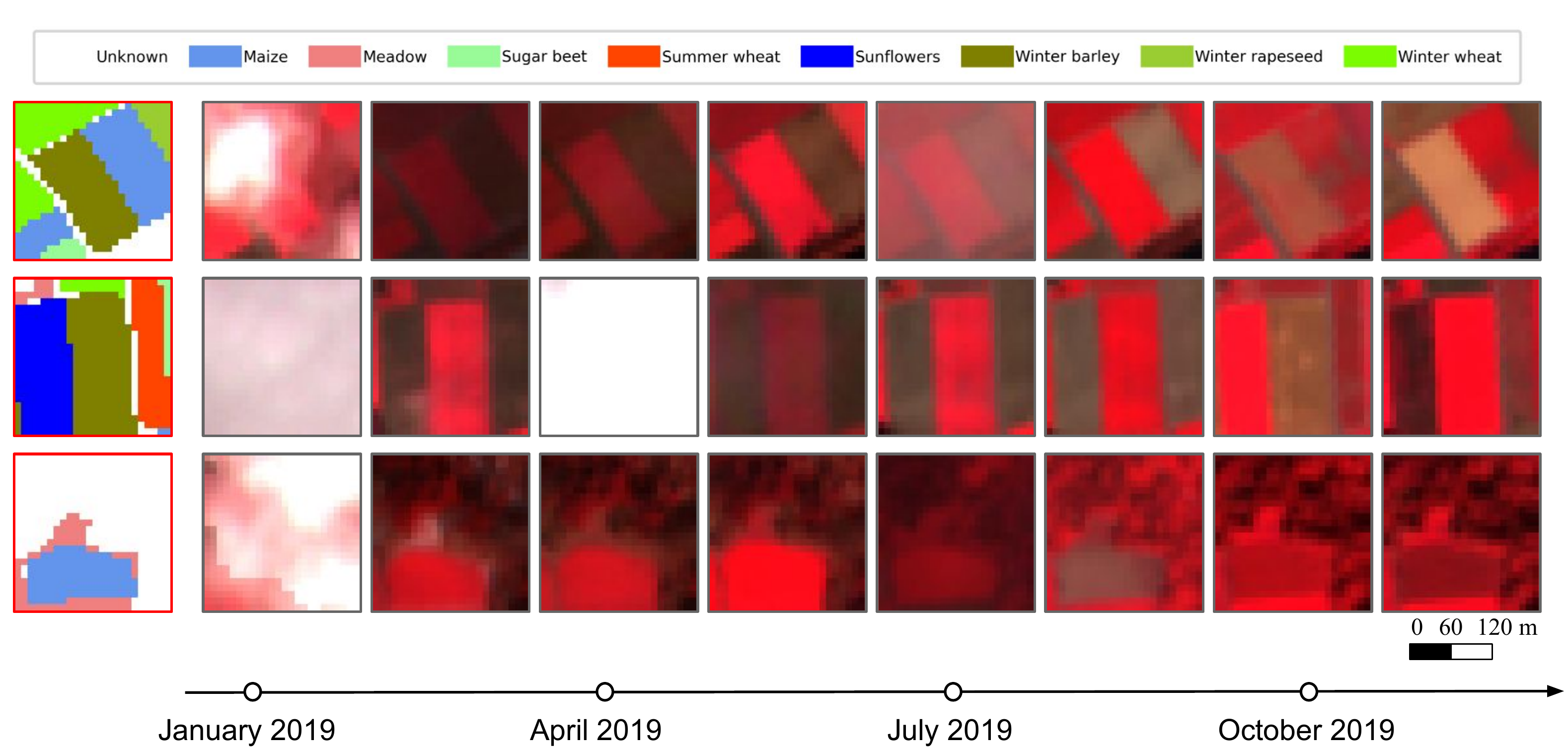}
    \caption{Example Sentinel-2 satellite images of the \emph{ZueriCrop} dataset. Each row shows randomly sampled images (false color composite: NIR-Red-Green) from a satellite image time-series. The first column shows the ground-truth where different colors correspond to different crop types.}
    \label{fig:data_example}
\end{figure}
\begin{figure}[t]
    \centering
        \includegraphics[width=1.\columnwidth]{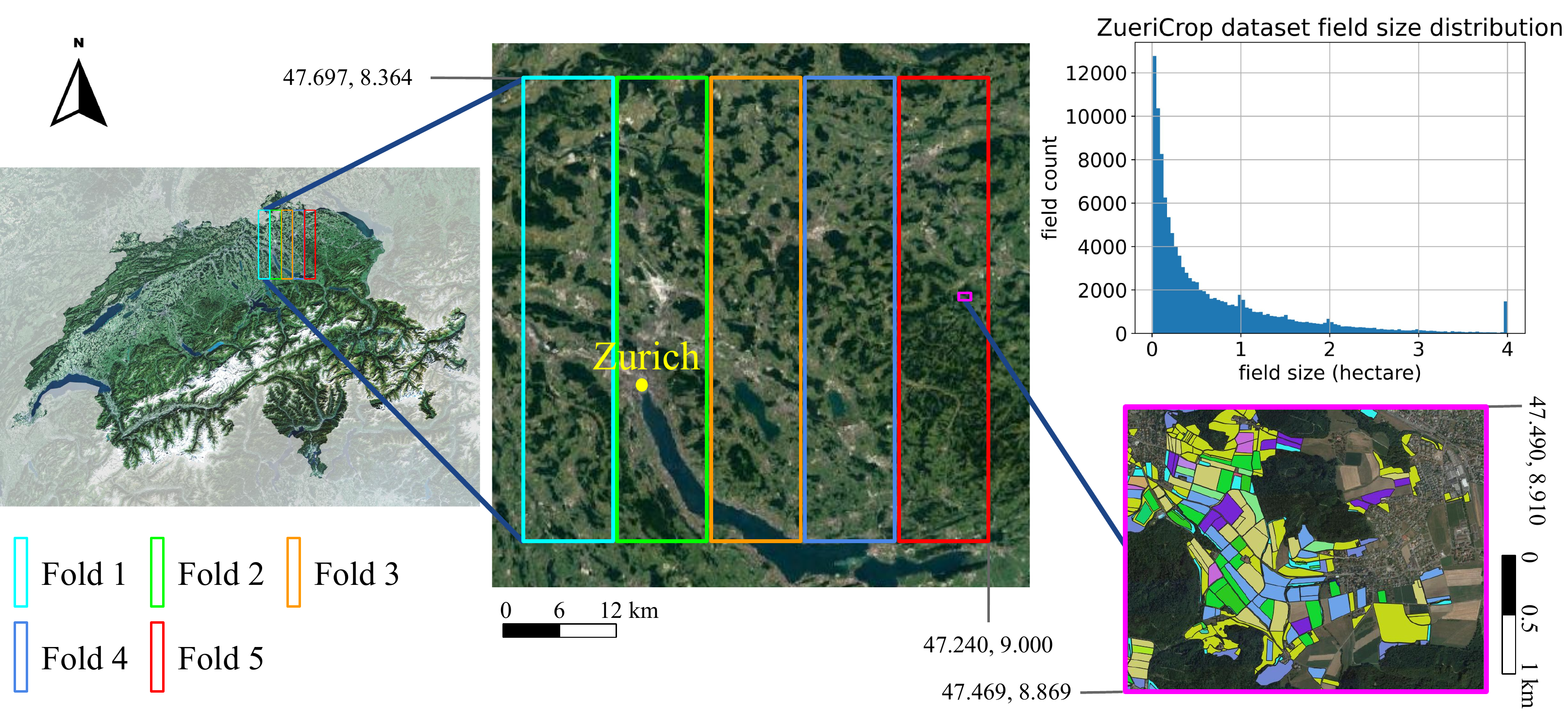}
    \caption{Overview of the \emph{ZueriCrop} dataset collected in 2019: Location inside Switzerland (left), Sentinel-2 image of the area of interest, overlaid with the geographical split used for cross-validation (center), example of GIS reference data for the main crop per field (bottom right), and distribution of field sizes (fields $>4$ hectares are pooled into one bin for visualization). The average field size is $0.72$ hectares.}
    \label{fig:data_region}
\end{figure}

\label{dataset} 
The \emph{ZueriCrop} dataset contains ground truth labels of 116,000 field instances. Each field instance consists of a polygon representing the borders of the field, and its dominant crop label in 2019. The ground truth labels of all 48 crop classes are provided by the Swiss Federal Office for Agriculture (FOAG) and correspond to the primary crop grown per field during the year. No information is provided about intermediate or cover crops, i.e., crops planted after harvesting the primary crop to cover the field over the winter in order to improve soil fertility and reduce disease pressure.
The input data is a time series of 71 multi-spectral Sentinel-2 Level-2A bottom-of-atmosphere reflectance images with a ground sampling distance (GSD) of 10 meters (Fig.~\ref{fig:data_example}). All input images are atmospherically corrected using the Sen2Cor v2.8 software package. The dataset is collected over a 50 km $\times$ 48 km area (Fig.~\ref{fig:data_region}) in the Swiss Cantons of Zurich and Thurgau between January 2019 and December 2019.

We subdivide the entire scene into smaller patches of $24~\text{px}\times24~\text{px}$. Patches without any ground-truth information are discarded. In the remaining patches the fraction of pixels without reference label is $\approx$48\%. Only those four spectral channels available at the highest, 10 m resolution (Red, Green, Blue, and Near-Infrared) are used because we observed that adding more channels did not significantly improve performance while increasing the computational cost, we refer the reader to~\ref{sec:num_input_ch} for an empirical evaluation how additional bands impact the model performance. We do not use any cloud detection method to discard patches with a high cloud cover because RNN architectures are robust to uninformative inputs. See Section~\ref{sec:cloud}.

\begin{figure}[t]
    \centering
        \includegraphics[width=1.\columnwidth]{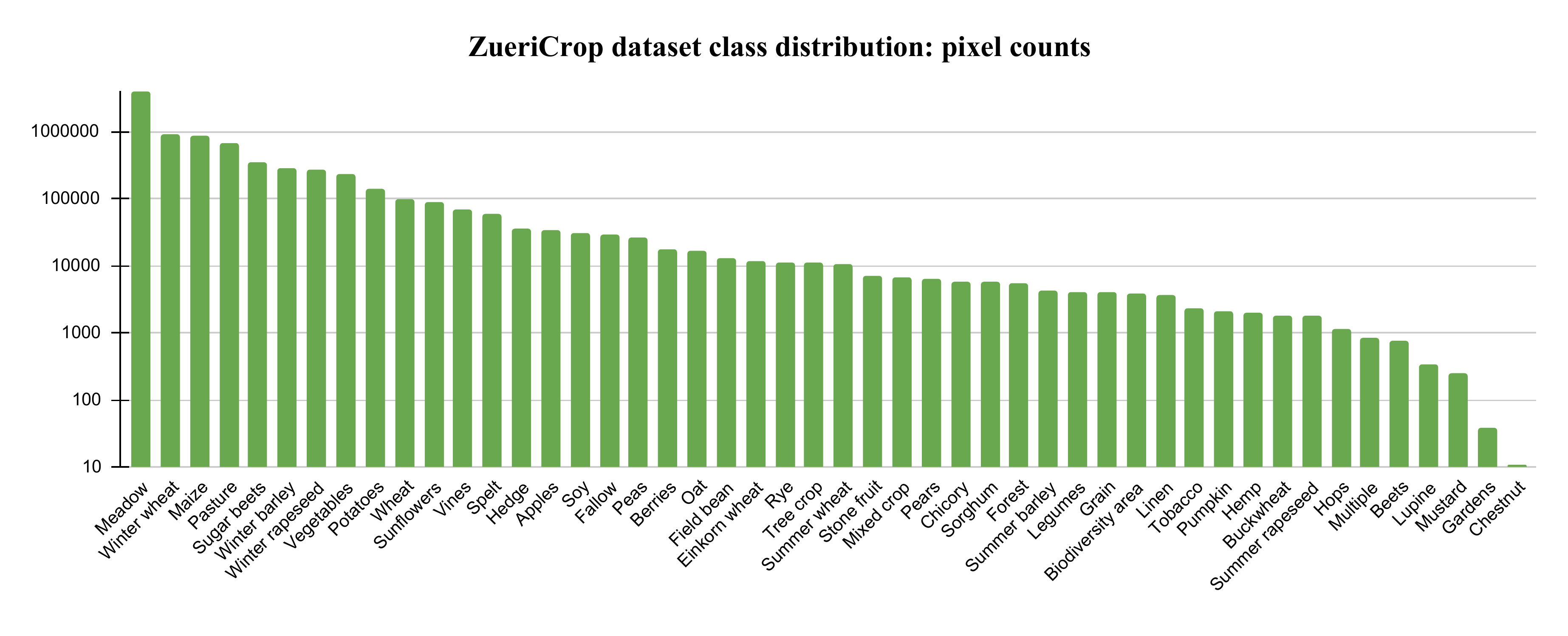}
    \caption{Class distribution in the \emph{ZueriCrop} dataset: Histogram of pixels. Note the logarithmic scale of the $y$-axis.} 
    \label{fig:class_dist}
\end{figure}

Switzerland has a small-structured agricultural system, where farmers are not allowed to grow crop after crop, but are required to adhere to a diverse crop rotation scheme. The average farm and field sizes in Switzerland are $21$ hectares and $1.5$ hectares, respectively with approximately $70\%$ of all Swiss agricultural land being grassland~\citep{BundesamtfurStatistik2020}, of which $12\%$ are temporary grasslands used in rotation with other crops~\citep{Stumpf2020}. This situation leads to a diverse set of crop classes with a highly skewed, imbalanced class distribution (Fig.~\ref{fig:class_dist}).

\subsection{Crop class hierarchy}\label{subsec:crophierarchy} 

We organize the 48 crop classes of the \emph{ZueriCrop} dataset into a hierarchy based on expert knowledge about the Swiss agricultural system (Fig.~\ref{fig:label_hier}).
%

The \nth{1} level of the class hierarchy was chosen with two goals in mind: \emph{(i)} separate the main categories found in the Swiss agricultural landscape and \emph{(ii)} group crop types according to their visual appearance in satellite images. For example, \emph{field crops} are grown in crop rows on fields, a feature that can be picked up by remote sensing. Another example is \emph{grassland}, the largest class in the dataset, which is a very heterogeneous class containing many different grassland types, mixtures and other land use scenarios. Their main feature is that they are permanently green with high biomass and generally low plant height. Other classes on the \nth{1} level are: \emph{orchards}, \emph{special crops} and \emph{forest}. Orchards can usually be identified by their specific planting patterns. 
Special crops are a pool of marginally grown specialist crops (e.g., asparagus, different berries and herbs) and have a very diverse set of sub-classes.

The \nth{2} level of the class hierarchy contains more refined versions of the preceding classes. Classes in the second level were selected to represent the plant family and agronomic use and practices in cultivation.
All crops of the \emph{small grain cereal} class are cultivated in a very similar manner in rows with little row spacing, similar plant seed density per square meter and similar erectile canopy structure with ears appearing at the top of the plant habitus (small grain as fruits). Similarly, the \emph{broad leaf row crop} class contains dicotyledonous plant species that are cultivated in rows and possess, in contrast to grain crops, a horizontal leaf surface pattern. 

The \nth{3} level distinguishes different crop species and is the finest level in our hierarchy. For some crops the ground truth was not reported at that granularity. In such cases the \nth{2}-level labels were copied. For the \emph{forest} class, \nth{1} level label is copied to \nth{2} and \nth{3} levels.


%

%% file: 04_experiments.tex
%
We compare the performance of the proposed hierarchical scheme to several baselines, and to competing state-of-the-art methods. In order to avoid any biases due to regional differences within the area of interest, we always perform 5-fold leave-one-out cross-validation and average the performance metrics across all five train/test splits. 
We divide the dataset into $5$ geographically disjoint strips of equal size as shown in Fig.~\ref{fig:data_region}, and use 4 strips as training set and the \nth{5} one as test set. In the ground truth, a single label is assigned to each field polygon, i.e., differences between administrative field boundaries and actual crop boundaries are not represented. To match that format, we assign the same label to all pixels within a field polygon by majority voting (See Appendix for results without majority voting).
Classification performance is evaluated pixel-wise, with four different metrics: overall accuracy, average per-class precision, average per-class recall, and average per-class F1-score (Table~\ref{table:result}).
Note that overall accuracy, unlike the other metrics, does not compensate for class frequencies. It measures correctness \emph{per pixel}, respectively area, but not correctness \emph{per label}, respectively crop type; and is thus dominated by the performance for (few) frequent crops.
%
On the contrary, the other metrics are better indicators of class-wise performance, as they are computed separately per class (independent of the absolute pixel count) and averaged.

As baseline for the hierarchical approach we also run the standard (single-level) convSTAR network~\citep{star} without intermediate outputs, losses, and label refinement. Hyper-parameters, like network depth, hidden state size and learning rate schedule are all set the same for the hierarchical method.
As further baseline experiments we add data augmentation and class-balanced losses to the standard convSTAR network, which are alternative, widely used techniques to improve the performance on imbalanced datasets. When balancing the loss, the contribution of a training example is weighted inversely proportional to the class frequency, such that every class (in principle) contributes an equal share of the total loss.
For data augmentation, training patches are sampled inversely proportional to the class frequency to achieve the same effect.
The class frequencies are found by counting pixels over the entire training set.

The use of a class-balanced loss functions causes a subtle difference, as it affects the gradient magnitude reducing the overall learning rate, which can potentially harm the training. Therefore, we also test another variant, denoted as Class-balance loss-2, where, after including the weight of the loss function for each class we compute the median effective learning rate and set it to match the default learning rate.
%
Moreover, we also evaluate the recently proposed, more robust re-weighting scheme of~\citet{cb_beta}.
That state-of-the-art technique uses the effective number of samples for each class to re-balance the loss. As the method has an additional, empirically chosen hyper-parameter $\beta$, we test different settings $\beta \in \{0.99, 0.999, 0.9999\}$.

The baselines described so far serve to isolate the impact of the proposed hierarchical labeling scheme, and to that end use the same convSTAR backbone. 
As an additional, external reference we also compare against Random Forest and other state-of-the-art deep learning methods that have been developed for multi-temporal crop classification.~\citet{marc_lstm} use LSTM,~\citet{tempCONV} propose a temporal convolutional network, and~\citet{breizhcrops,russwurm2020self} design a Transformer network. All four methods process  pixels individually, unlike our approach that aggregates context information from image patches with convolutions.
Convolutions and recurrence are used separately by~\citet{fcn}, who extract features from the images with a convolutional architecture in the style of U-Net~\citep{unet} and apply convolutional LSTM  to the resulting feature vectors to represent their temporal evolution.
Also, U-Net~\citep{unet} itself is compared since it has become a standard architecture for image segmentation and has been used in remote sensing (e.g., \cite{stoian2019land, flood2019using}). Early temporal fusion, spectral channels of different timestamps are concatenated in the channel direction, is applied to deal with multi-temporal data.
An integrated, convolutional variant of GRU is already used by~\citet{bigru}. This work is, from a technical point of view, similar to our convSTAR baseline, except that it uses a different form of recurrence and employs a bi-directional approach.

%% file: 05_result.tex

In this section, we first compare the performance of our proposed ms-convSTAR against baseline methods (Table~\ref{table:result}), as well as other state-of-the-art methods (Table~\ref{table:result_sota}) on the \emph{ZueriCrop} dataset. In the ablation study (Section~\ref{subsec:ablation}), we evaluate the effectiveness of our label refinement component (Table~\ref{table:result}), and compare different convRNN types (Table~\ref{table:ablation_RNN}). 
We also discuss how the proposed method can be used in combination with the hierarchical label tree to provide more certain prediction by adjusting the label coarseness at prediction time.
Finally, we also discuss the robustness of ms-convSTAR against clouds.

\begin{table}[t]
\begin{center}
\begin{tabular}{l|ccc|c}
\toprule
{\bf Method} &{\bf Prec (\%)} &{\bf Rec (\%)} &{\bf F1 (\%)} & {\bf Acc (\%)}\\
\midrule
{convSTAR}      &{40.2} &{37.3} &{37.2} &{87.3}\\
%
{\hspace{4mm}+ Data augmentation}      &{48.3} &{39.3} &{41.1} &{85.0}\\
{\hspace{4mm}+ Class-balanced loss}      &{26.9} &{32.7} &{28.2} &{75.6}\\
{\hspace{4mm}+ Class-balanced loss-2}      &{26.5} &{31.7} &{27.3} &{74.0}\\
{\hspace{4mm}+ Cui 2019, $\beta=0.99$}      &{42.2} &{37.5} &{36.5} &{87.3}\\
{\hspace{4mm}+ Cui 2019, $\beta=0.999$}     &{39.4} &{35.7} &{35.6} &{87.4}\\
{\hspace{4mm}+ Cui 2019, $\beta=0.9999$}     &{43.3} &{39.1} &{39.4} &{87.1}\\
%
{ms-convSTAR w/o LR}  &{59.3} &{47.1} &{49.8}&{87.6}\\
{ms-convSTAR}  &{\textbf{60.1}} &{\textbf{49.8}} &{\textbf{52.4}}&{\textbf{88.0}}
\\ \bottomrule 
\end{tabular}
\end{center}
\caption{Performance comparison between ms-convSTAR (bottom row) and  non-hierarchical baseline methods. We compare against standard (1-level) convSTAR (top row) as well as further baselines that extend convSTAR with different techniques intended to compensate class imbalance. Precision, recall and F1-score are mean values over all classes. All numbers are averaged over 5 cross-validation folds. The best score for each metric is shown with \textbf{bold}.} 
\label{table:result}
\end{table}

\begin{table}[t]
\begin{center}
\tabcolsep=0.01cm
\begin{tabular}{l|ccc|c}
\toprule
{\bf Method}  &{\bf Prec(\%)} &{\bf Rec(\%)} &{\bf F1(\%)}&{\bf Acc(\%)}  \\
\midrule
{Random Forest*}     &{46.4} &{\underline{40.7}} &{38.9} &{78.8}  \\
{LSTM \citep{marc_lstm}}     &{37.7} &{27.9} &{29.2} &{84.1}  \\
{TCN \citep{tempCONV}}     &{39.2} &{27.7} &{29.3} &{83.5}  \\
{Transformer \citep{breizhcrops}}     &{\underline{56.8}} &{38.4} &{42.3} &{85.4}  \\
{2D-CNN (U-Net)}     &{34.6} &{25.7} &{26.7} &{82.2}  \\
{U-Net+convLSTM \citep{fcn}}     &{47.7} &{32.8} &{35.2} &{85.0} \\
{Bi-convGRU \citep{bigru}}      &{55.0} &{39.6} &{\underline{42.5}} &{\underline{86.4}} \\
\midrule
{ms-convSTAR}  &{\textbf{60.1}} &{\textbf{49.8}} &{\textbf{52.4}}&{\textbf{88.0}} 
\\ \bottomrule 
\end{tabular} 
\end{center}
\caption{Performance comparison of ms-convSTAR (bottom row) with state-of-the-art methods. Precision, recall and F1-score are mean values over all classes. All numbers are averaged over 5 cross-validation folds. The best score for each metric is shown with \textbf{bold} and the second best is \underline{underlined}. *Random Forest is trained with a balanced dataset by under-sampling the majority classes which leads to improved class-wise performance.} 
\label{table:result_sota}
\end{table}

\subsection{Performance Comparison}


%
We find that among the baselines, simple data augmentation performs best, but all baselines are clearly outperformed by the proposed, hierarchical ms-convSTAR, on all performance metrics. See Table~\ref{table:result}.
Most significantly, there are significant improvements on those metrics that compensate for class frequencies and average measure per-class performance. Our proposed ms-convSTAR increases mean class precision by $>11$ percentage points, and mean class recall by $>10$ percentage points. Accordingly, the F1-score (their harmonic mean) increases by $>11$ percentage points. These results indicate that our method improves in particular the classification of less frequent classes, which was the initial motivation for using the crop label hierarchy and developing ms-convSTAR. 
Data augmentation, where we over-sample the minority classes during training (i.e., patches with rare classes are sampled inversely proportional to their frequencies), improves the F1-score by 2.9 percentage points compared to the baseline convSTAR. However, it degrades the overall accuracy significantly by 2.3 percentage points, because over-sampling rare classes induces a global bias towards those classes and degrades performance for the dominant classes. 
Adding a standard class-balanced loss to the baseline convSTAR approach significantly decreases performance across all measures, whereas the state-of-the-art class-balanced loss technique~\citep{cb_beta} can improve the F1-score a little (by 2.2 percentage points with $\beta=0.9999$) but slightly reduces the overall accuracy (by $0.2$ percentage points). In summary, even though a class-balanced loss or data augmentation bring a mild  improvement for minority classes, the gains are not very big, and they tend to reduce the overall performance in return. In contrast, our proposed ms-convSTAR greatly boosts performance for mean precision, mean recall and F1-score, while improving overall accuracy, too.

\begin{figure}[t]
    \centering
        \includegraphics[width=1.\columnwidth]{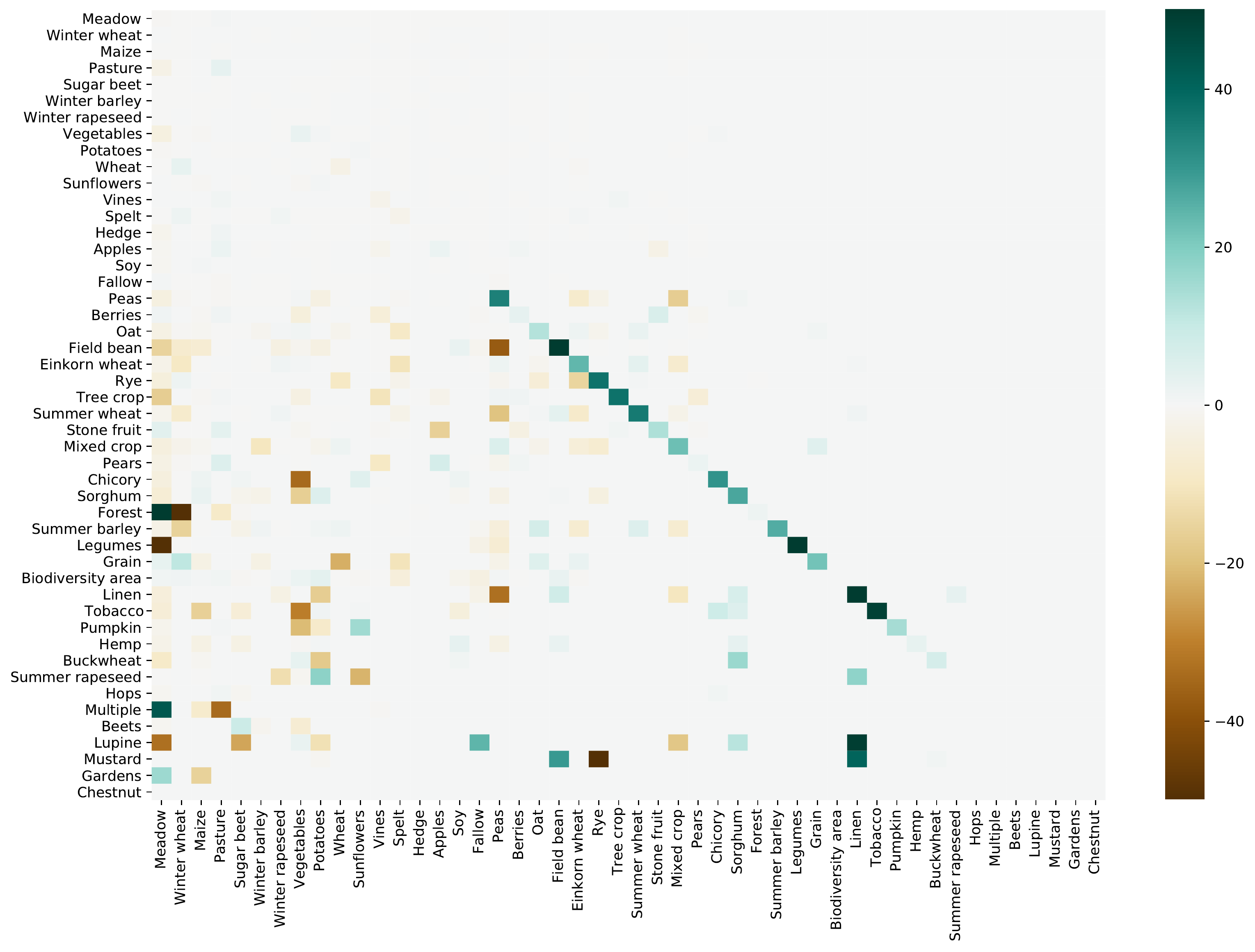}
    \caption{Benefit of ms-convSTAR: Difference between normalized confusion matrices. Averaged over 5 cross-validation folds. Green denotes margins in favour of ms-convSTAR (higher
correctness on the diagonal, respectively lower confusion off the diagonal), brown denotes margins in favour of the baseline counterpart (convSTAR).}
    \label{fig:cm_dif}
\end{figure}

\begin{figure}[t]
    \centering
        \includegraphics[width=1.\columnwidth]{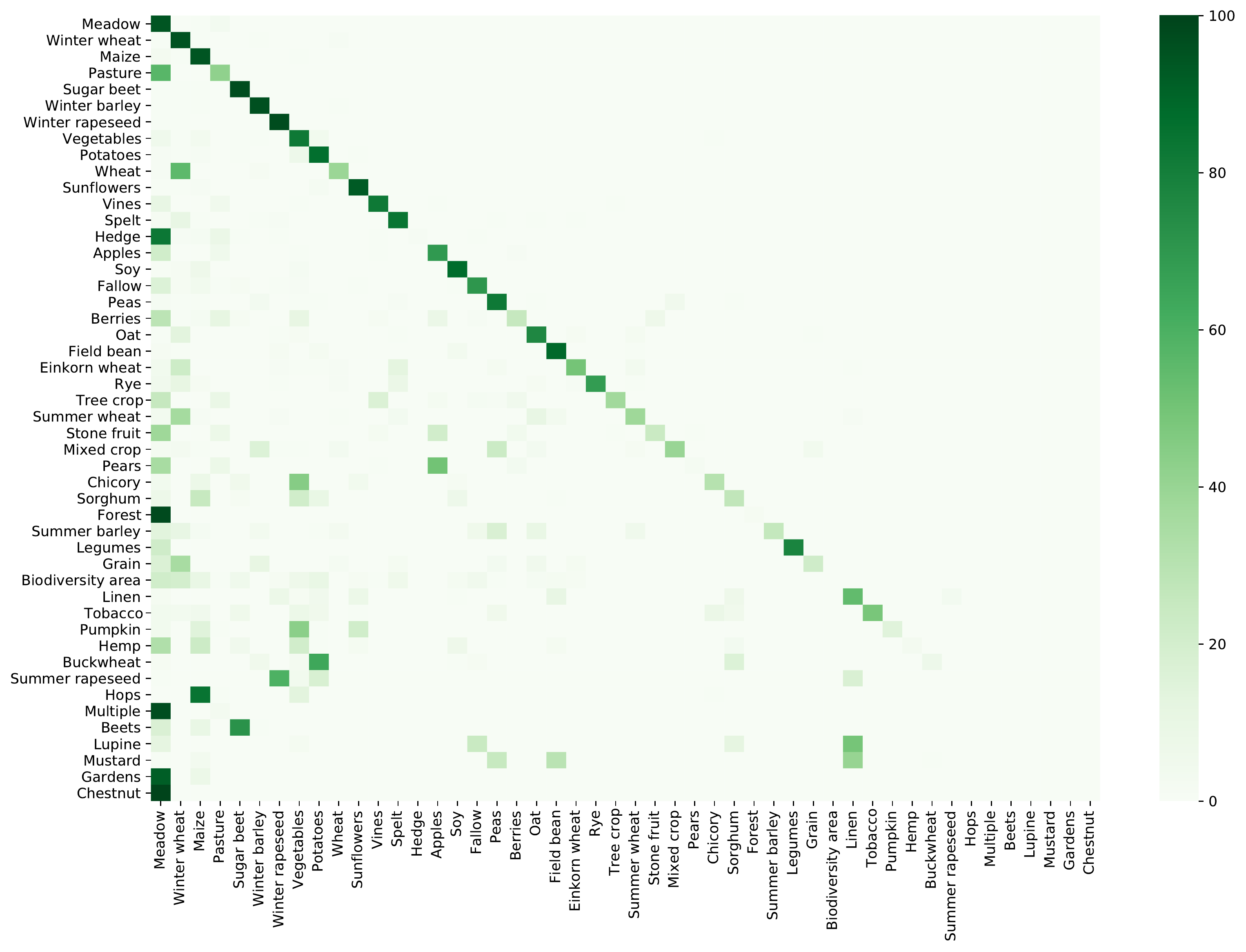}
    \caption{Confusion matrix for the proposed method. Averaged over 5 cross-validation folds. Rows show true labels and columns show predicted labels. The sum of each row is normalized to $1$.}
    \label{fig:cm}
\end{figure}

Fig.~\ref{fig:cm_dif} illustrates the difference between the confusion matrix of results achieved with proposed ms-convSTAR (Fig.~\ref{fig:cm}) and with its baseline counterpart convSTAR. ms-convSTAR improves the performance for many less frequent classes like \emph{stone fruit}, \emph{legumes}, \emph{tobacco} while it does not harm the performance for frequent classes like \emph{meadow} or \emph{maize}. %
We note that for exceedingly rare classes the performance does not improve. These classes have too few pixels (typically $<1000$) to be represented well, moreover they are often only present in some of the five cross-validation stripes, such that they do not appear in either the training set or the test set. For completeness, we nevertheless leave those classes in the dataset.

%
%

In Table~\ref{table:result_sota}, we compare the proposed ms-convSTAR (bottom row) to a number of state-of-the-art methods for crop classification from image time series. ms-convSTAR significantly improves performance across all measures. Again the gains are mostly due to better classification of rare classes, as indicated by an improvement of $>9.9$ percentage points in F1-score.
Model implementations are taken from official codebases, and similar hyperparameters with the proposed method are used as the hidden size, the initial learning rate, the number of epochs, etc.; for other model-specific hyperparameters (e.g., number of layers in TCN~\citep{tempCONV}, number of heads in Transformer~\citep{breizhcrops}), values from original papers are used. See Appendix for more details.
Random Forest parameter settings follow~\cite{russwurm2020self}. To achieve better classification performance for the Random Forest, we augmented the raw input reflectance with the NDVI and the training dataset is randomly under-sampled such a way that number of per class samples is limited to 10K ($\approx$ the median of the number of per class samples) like done in~\cite{russwurm2020self}.
%

\begin{figure}[t]
    \centering
        \includegraphics[width=1.\columnwidth]{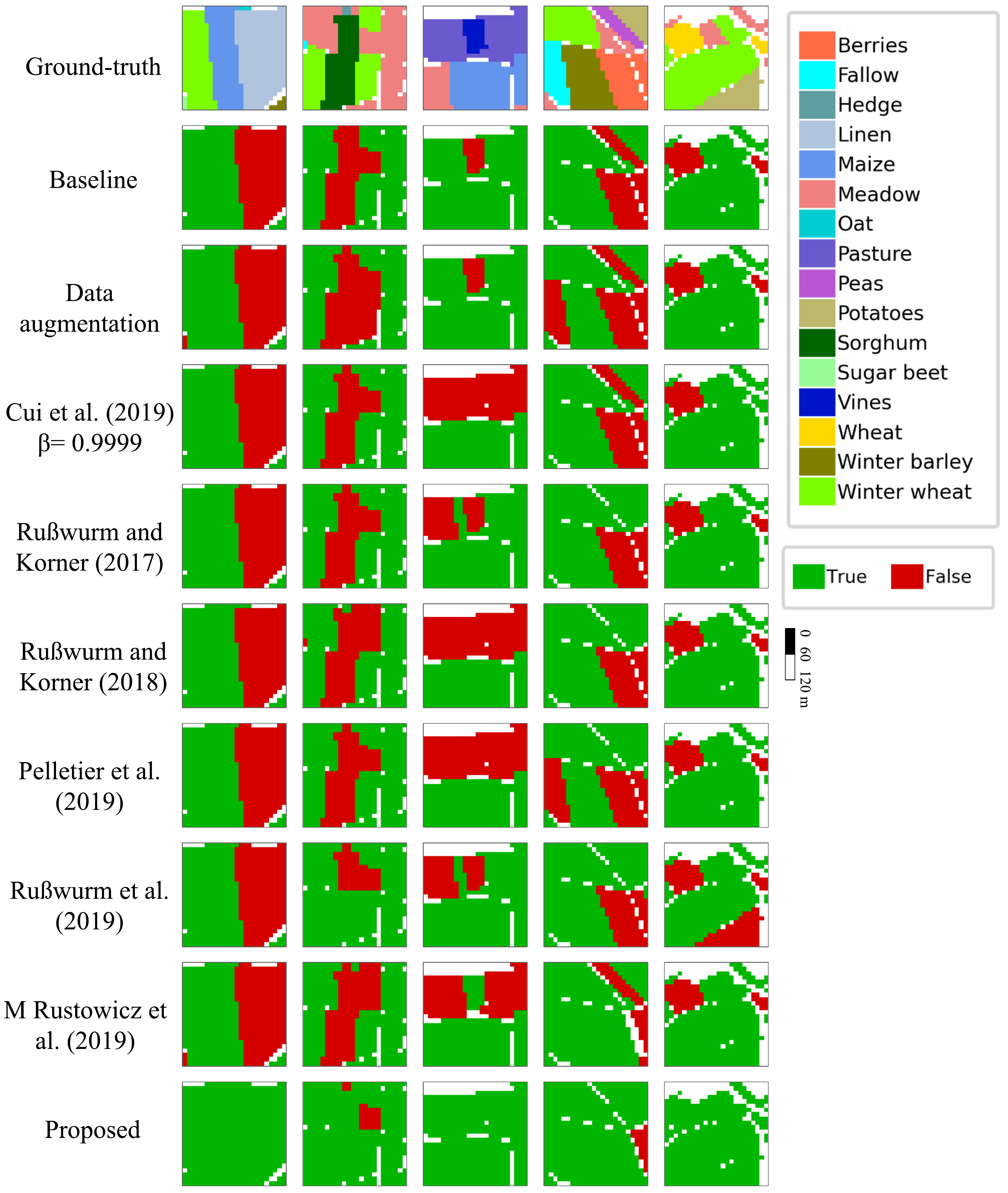}
    \caption{Visual results for five different samples (train-test fold 2). Green color indicates correct classification, red are mis-classified pixels (respectively, fields).}
    \label{fig:qualitative_2}
\end{figure}

\begin{figure}[t]
    \centering
        \includegraphics[width=1.\columnwidth]{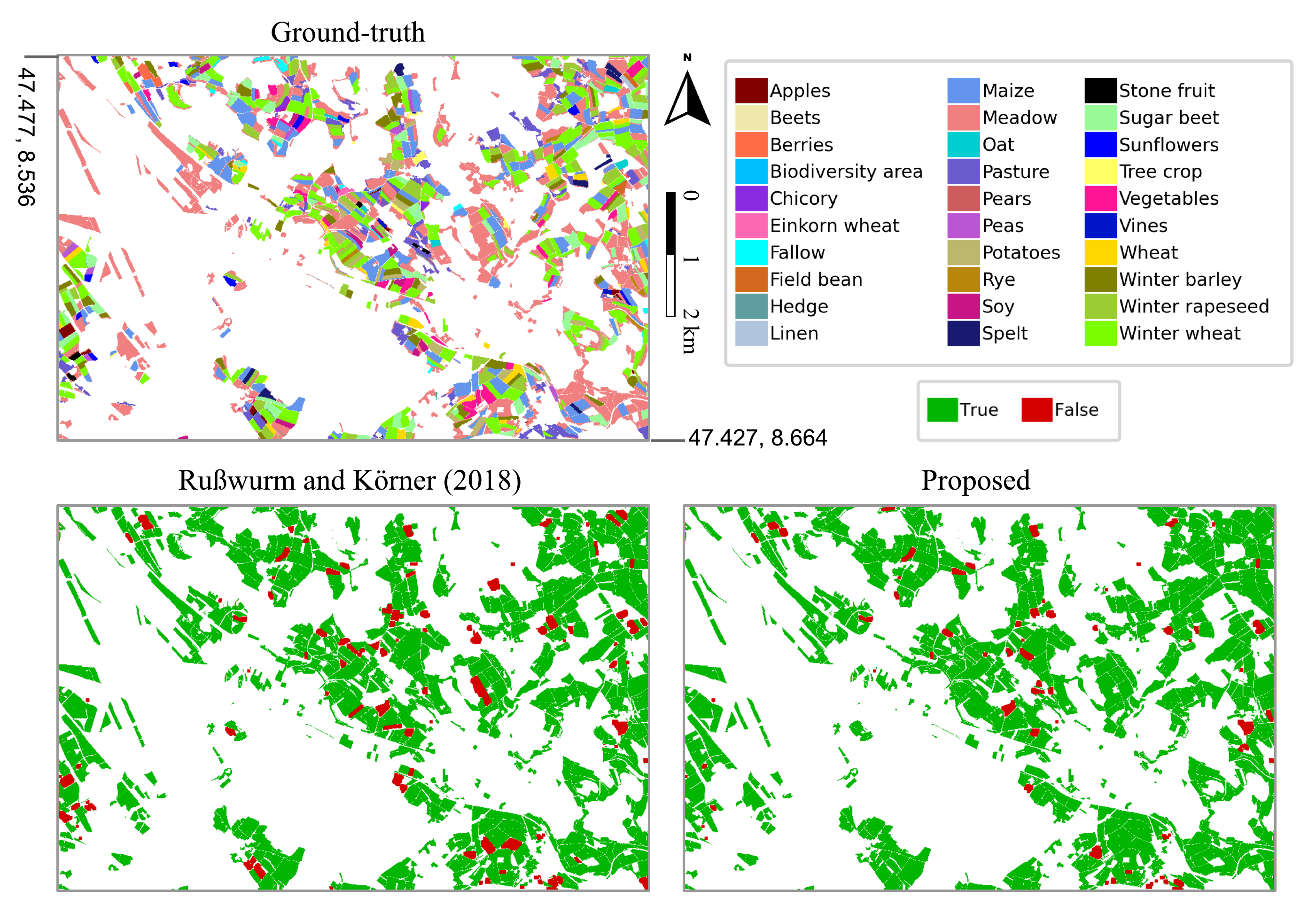}
    \caption{Qualitative comparison: ms-convSTAR vs.\ the best-performing alternative on the \emph{ZueriCrop} data.}
    \label{fig:qualitative}
\end{figure}

We show qualitative comparisons for several output samples in Fig.~\ref{fig:qualitative_2}. Our new method does what it is designed for: it correctly predicts rare classes like \emph{linen} or \emph{sorghum}, where all other methods fail. Another qualitative comparison to the closest competitor~\citep{bigru} is shown in Fig.~\ref{fig:qualitative} for a larger region.
We point out that both approaches, ms-convSTAR and~\citep{bigru}, typically mis-classify the same fields. But the number of mistakes is significantly smaller with ms-convSTAR. For qualitative results without polygon aggregation, e.g., for mapping in the absence of field boundaries, see Appendix (Fig.~\ref{fig:mapping}).
%
%

%

\begin{table}[t]
\begin{center}
\begin{tabular}{c|ccc|c}
\toprule
{\bf Method}  &{\bf Prec (\%)} &{\bf Rec (\%)} &{\bf F1 (\%)}&{\bf Acc (\%)}  \\
\midrule
\multirow{6}{*}{\rotatebox[origin=t]{90}{3 layers}}
{\hspace{6mm} convLSTM}    &{31.5} &{27.9} &{28.1} &{84.6}  \\
{\hspace{5mm} ms-convLSTM}      &{37.7} &{30.1} &{31.1} &{83.9}  \\
{\hspace{13mm} convGRU}      &{43.1} &{38.2} &{38.6} &{87.2} \\
{\hspace{7mm} ms-convGRU}      &{49.8} &{38.6} &{40.8} &{85.8}\\  
{\hspace{11mm} convSTAR}       &{48.3} &{42.6} &{43.5} &{\underline{87.8}} \\
{\hspace{5mm} ms-convSTAR}      &{\underline{54.6}} &{\underline{42.7}} &{\underline{45.3}} &{86.8}\\  
\midrule
\multirow{6}{*}{\rotatebox[origin=c]{90}{6 layers}}

{\hspace{6mm} convLSTM}       &{1.1} &{2.3} &{1.4} &{47.2} \\

{\hspace{5mm} ms-convLSTM}       &{40.5} &{33.6} &{34.7} &{85.2}\\  
{\hspace{13mm} convGRU}       &{15.8} &{15.7} &{15.0} &{71.1} \\
{\hspace{7mm}  ms-convGRU}    &{52.2} &{42.9} &{44.7} &{86.9}\\  

{\hspace{11mm} convSTAR}     &{40.2} &{37.3} &{37.2} &{87.3}\\
{\hspace{5mm} ms-convSTAR}  &{\textbf{60.1}} &{\textbf{49.8}} &{\textbf{52.4}}&{\textbf{88.0}}

\\ \bottomrule 
\end{tabular}
\end{center}
\caption{Performance comparison of multi-stage convRNNs (ms-convRNNs). All numbers are averaged over 5 cross-validation folds. The best score for each metric is shown with \textbf{bold} and the second best is \underline{underlined}. }
\label{table:ablation_RNN}
\end{table}

\subsection{Ablation Study}\label{subsec:ablation}
As an ablation study, we first evaluate the performance gain due to the label refinement module by also running our hierarchical ms-convSTAR model without the label refinement module (denoted as ms-convSTAR w/o LR). 
Results in Table~\ref{table:result} show that the CNN-based label refinement consistently improves all performance metrics, in particular per-class performance.

We then investigate the importance of the RNN cell type used in the network (Table~\ref{table:ablation_RNN}). We construct the hierarchical multi-stage network (ms-convX) and its non-hierarchical baseline counterparts (convX) using the most popular RNN cells: LSTM and GRU. 
%
%
%
%
We experimentally evaluate 3-layer and 6-layer convRNN versions and corresponding ms-convRNN versions (Table~\ref{table:ablation_RNN}). 
%
While convLSTM and convGRU produce acceptable results for the 3-layer version without hierarchy, they perform very poorly for the deeper, 6-layer versions. As we have shown in \citep{star}, both cell types LSTM and GRU suffer from gradient vanishing problems, which get more severe with deeper architectures. Using our proposed hierarchical approach fixes these gradient problems and leads to substantial improvements with respect to their baseline versions without hierarchy. 
Unlike convLSTM and convGRU, convSTAR does not suffer from gradient issues at any point and performs well for all settings, outperforming all existing approaches in overall accuracy (Table~\ref{table:result_sota}). Combining convSTAR with the proposed hierarchical approach leads to superior performance with a clear margin above all other methods on the \emph{ZueriCrop} dataset.
%

\subsection{Simultaneous multi-level classification}
\label{sec:simultaneous}
%
\begin{figure}[t]
    \centering
    \begin{subfigure}[b]{0.49\textwidth}
        \includegraphics[width=\textwidth]{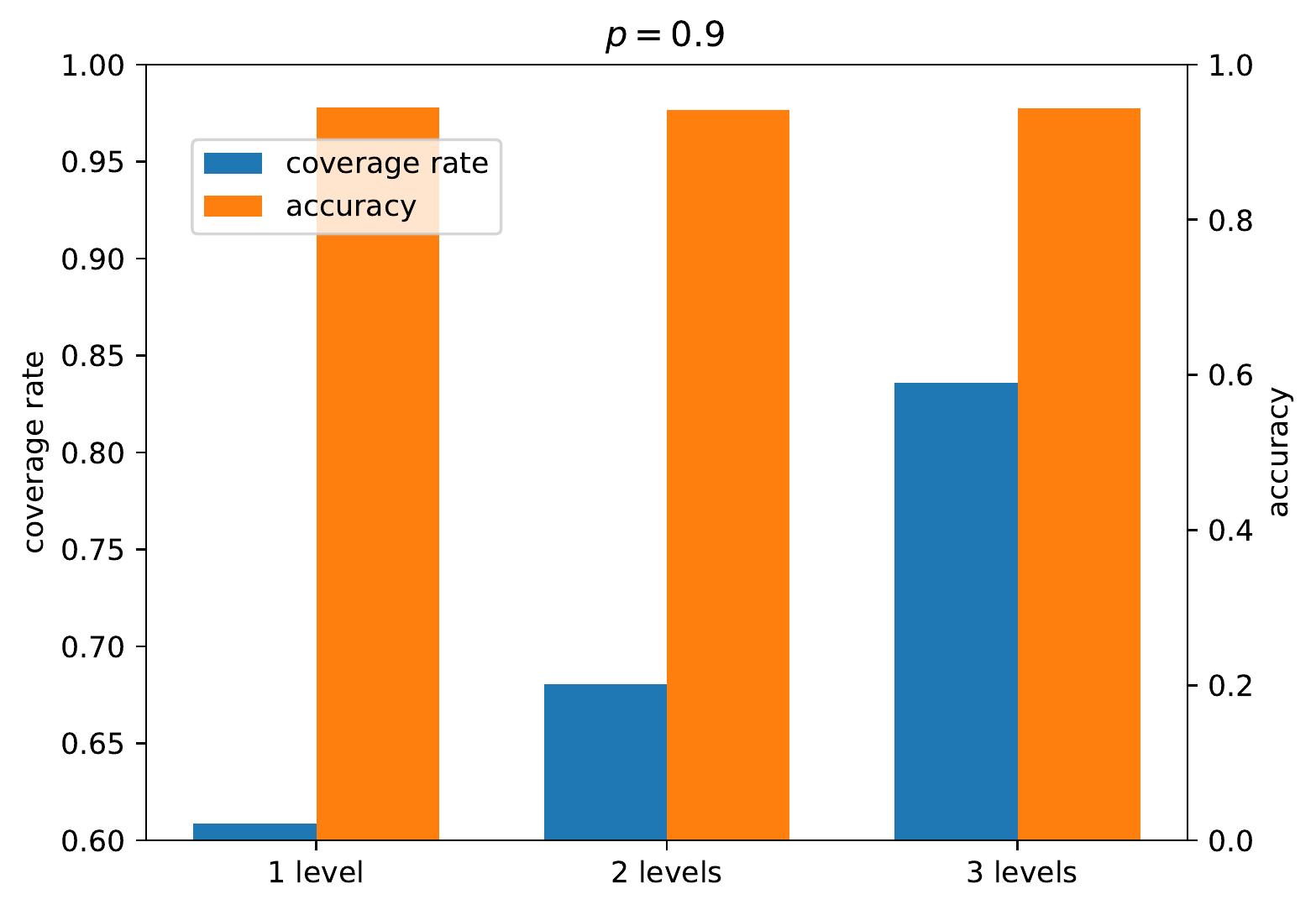}
        \caption{$p=0.9$}
    \end{subfigure}  
    \begin{subfigure}[b]{0.49\textwidth}
        \includegraphics[width=\textwidth]{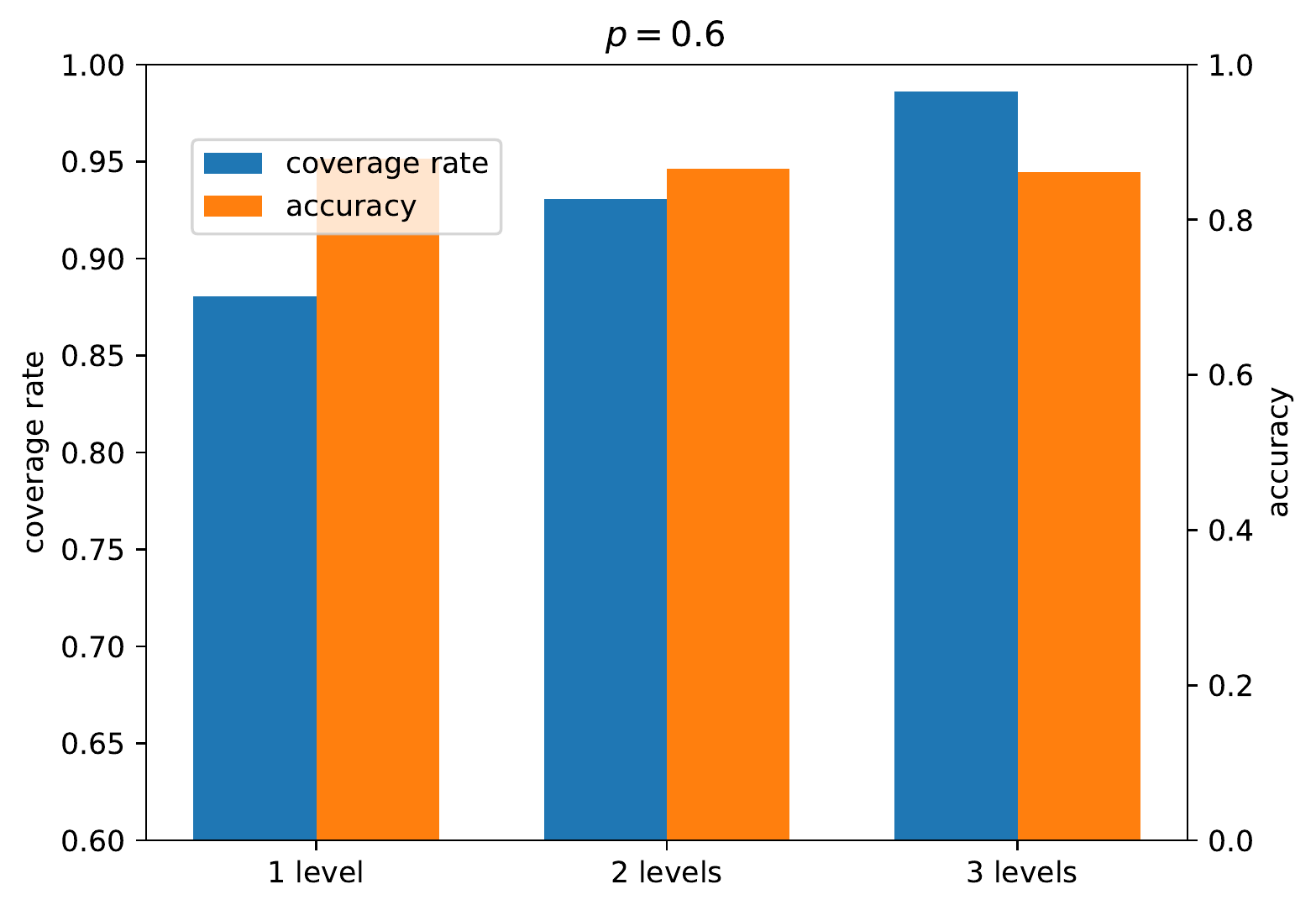}
        \caption{$p=0.6$}
    \end{subfigure}  
    \caption{Performance of models with different numbers of hierarchical levels at fixed confidence values: (a) $p=0.9$, (b) $p=0.6$. The fraction of classified pixels (coverage rate, blue) grows with an increasing number $N$ of hierarchy levels (computed with a single train-test fold). See Section~\ref{sec:simultaneous}.}\label{fig:coverage2}
\end{figure}
\begin{figure}[t]
    \centering
    \begin{subfigure}[b]{0.49\textwidth}
        \includegraphics[width=\textwidth]{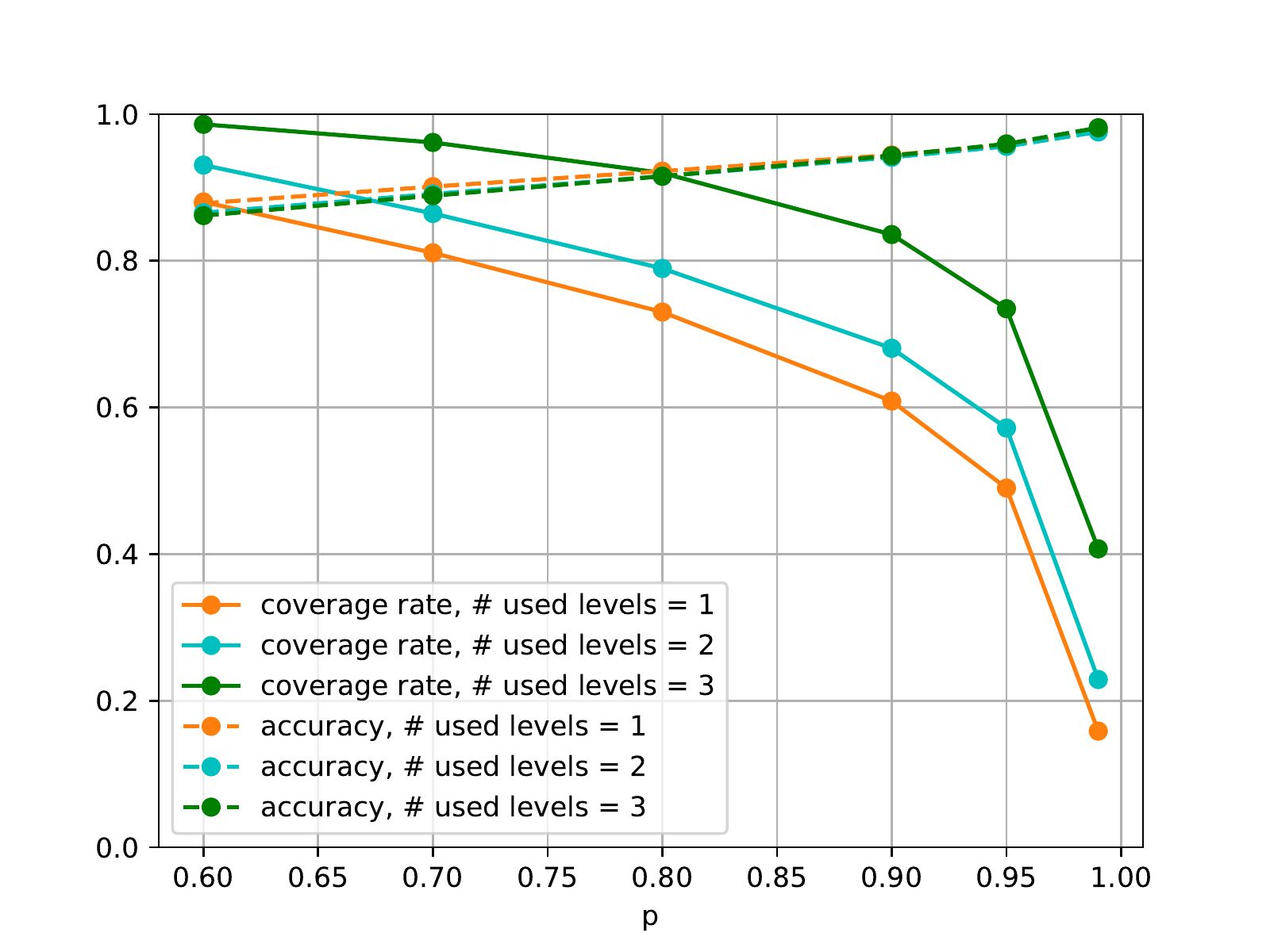}
        \caption{}
        \label{fig:coverage_a}
    \end{subfigure}  
    \begin{subfigure}[b]{0.49\textwidth}
        \includegraphics[width=\textwidth]{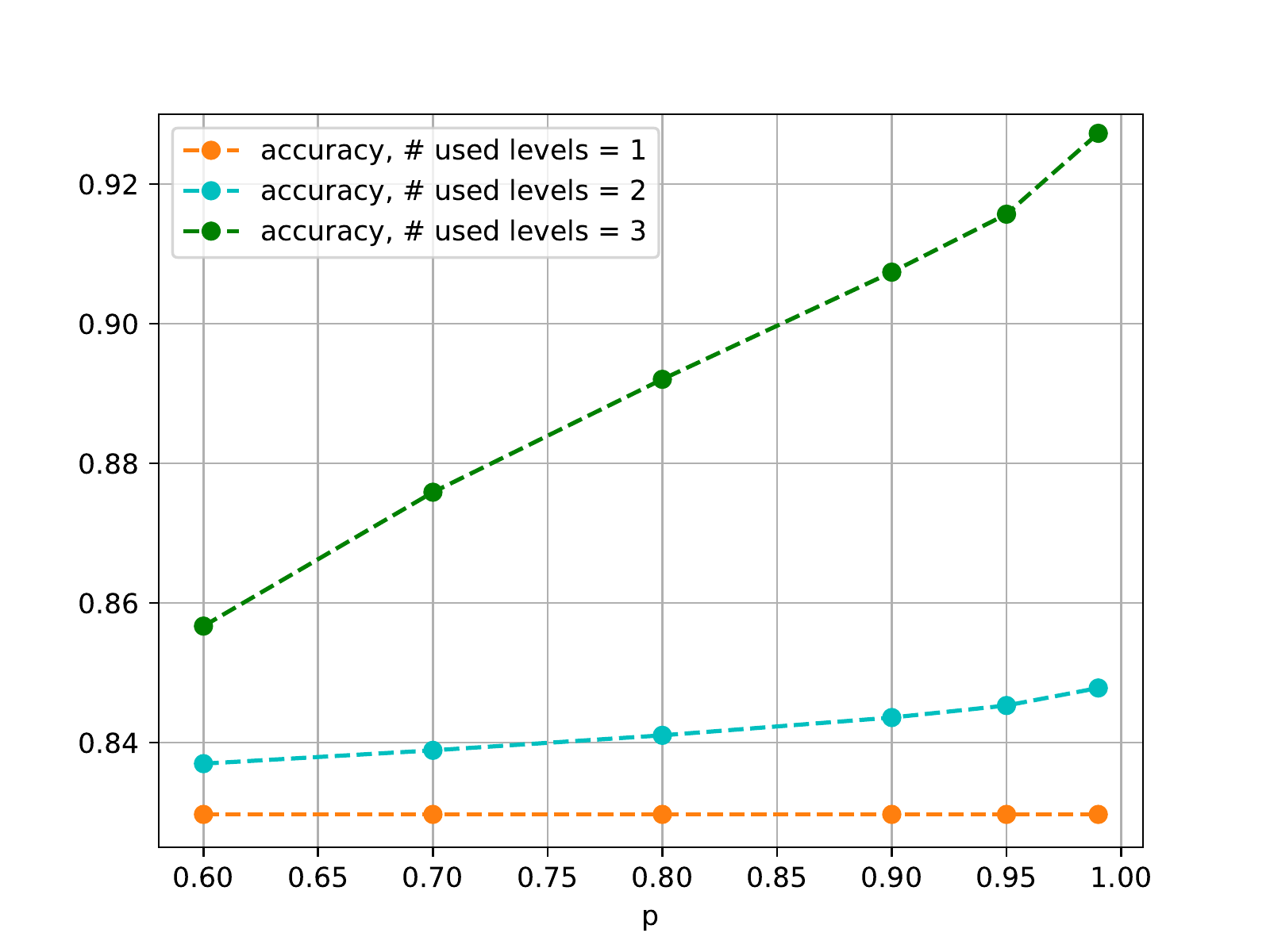}
        \caption{}
        \label{fig:coverage_b}
    \end{subfigure}  
    \caption{Performance of models with different numbers of hierarchical levels: (a) Coverage rate and model performance vs.\ confidence value $p$, (b) performance vs.\ confidence value $p$ in full coverage case for models with different number of hierarchy levels (computed with a single train-test fold). See Section~\ref{sec:simultaneous}.}\label{fig:coverage}
\end{figure}
%

An attractive property of ms-convSTAR is that it simultaneously predicts per-pixel labels at three different levels in the class hierarchy. 
Performance of proposed ms-convSTAR at different levels of granularity is given in Table~\ref{table:result_level}. As we expected, the accuracy of the model improves while going to the coarser levels.

\begin{table}[t]
\begin{center}
\begin{tabular}{l|ccc|c}
\toprule
{\bf Level} &{\bf Prec (\%)} &{\bf Rec (\%)} &{\bf F1 (\%)} & {\bf Acc (\%)}\\
\midrule
{\nth{1}}  &{80.3} &{52.9} &{57.0} &{96.3}\\
{\nth{2}}  &{73.0} &{51.0} &{54.5} &{89.2}\\
{\nth{3}}  &{60.1} &{49.8} &{52.4} &{88.0}
\\ \bottomrule 
\end{tabular}
\end{center}
\caption{Performance of proposed ms-convSTAR at different levels of granularity. All numbers are averaged over 5 cross-validation folds.} 
\label{table:result_level}
\end{table}

Such feature makes also possible to choose the granularity of the labels according to the output (confidence) scores. In this way one can produce reliable maps where most pixels are assigned a label and all labels have sufficiently high confidence -- at the cost of only assigning a coarse-grained label to some pixels were the fine-grained answer is too uncertain.
For instance if the model is uncertain about deciding for either \emph{apple} or \emph{pear} orchard, the coarser label \emph{orchard} can be assigned with much higher confidence. In a number of applications this coarser level of annotation is good enough. An example are the summary statistics computed by the Swiss federal administration, where a large degree of coverage with coarse labels is critical.
Moreover, coarse but correct and (nearly) complete answers are a lot more useful for downstream GIS processing: for instance, mapping orchards is much easier if the user knows which polygons have not yet received a fine-grained label and must be checked.

To quantify this effect, we measure the area that is classified above a given confidence score, and compare ms-convSTAR models with different numbers of hierarchy levels.
We set a prediction confidence value $p$, and, for each pixel switch to the next-coarser label if the confidence level $p$ is not achieved.
Fig.~\ref{fig:coverage2} shows the overall accuracy and the coverage rate (the proportion of pixels with a confident label at any granularity) for two confidence levels, $p=0.6$ and $p=0.9$. As can be seen if we wish to retain only confident predictions and set $p=0.9$, the 3-levels hierarchy brings a large benefit. While accuracy (computed over covered areas) stays the same, coverage improves by $\approx0.20-0.25$, although, obviously, some areas are only predicted at a coarse label resolution. 

%
Fig.~\ref{fig:coverage_a} shows how the coverage rate and the model performance change as functions of the confidence value $p$. One can see that, as expected, when decreasing $p$ the  coverage increases whereas the accuracy decreases. Fig.~\ref{fig:coverage_b} shows how the model performance improve with increasing $p$ in the full coverage case (every pixel is classified aat the coarsest available level regardless of the $p$ value). The full, 3-level model boosts performance significantly with increasing $p$; in contrast, a basic 1-level model does not show this behaviour. This analysis suggests that the imposed label hierarchy is actually meaningful for the problem, in the sense that the coarser classes are indeed easier to discriminate than their finer sub-classes.
%
Moreover, the results once more support our hierarchical multi-level scheme: additional hierarchy levels consistently help to predict more pixels correctly and confidently, for any value of $p$.
%



\subsection{Cloud Robustness}
\label{sec:cloud} 
\begin{figure}[t]
    \centering
    \begin{subfigure}[b]{0.49\textwidth}
        \includegraphics[width=\textwidth]{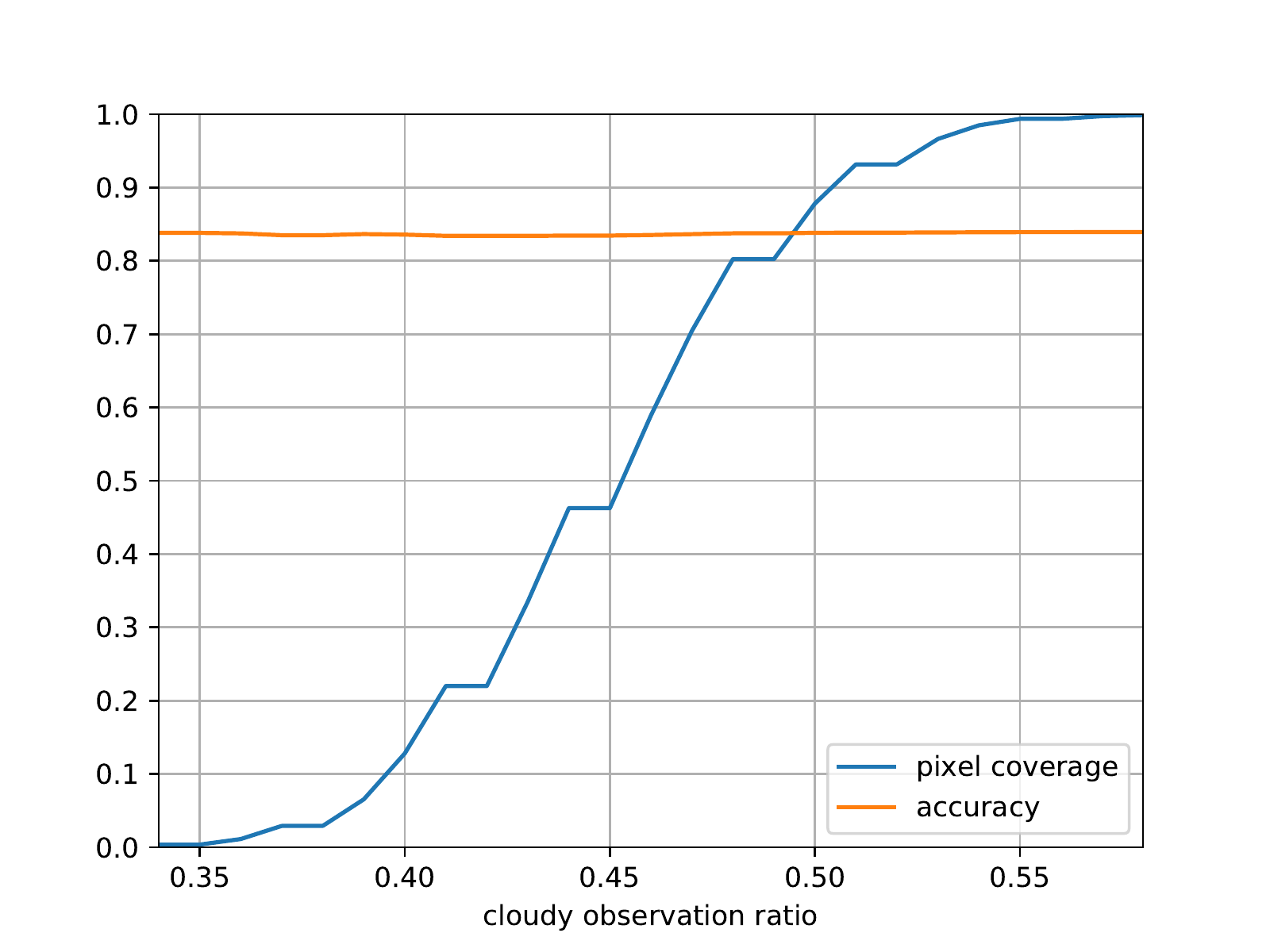}
        \caption{$c=0.1$}
    \end{subfigure}  
    \begin{subfigure}[b]{0.49\textwidth}
        \includegraphics[width=\textwidth]{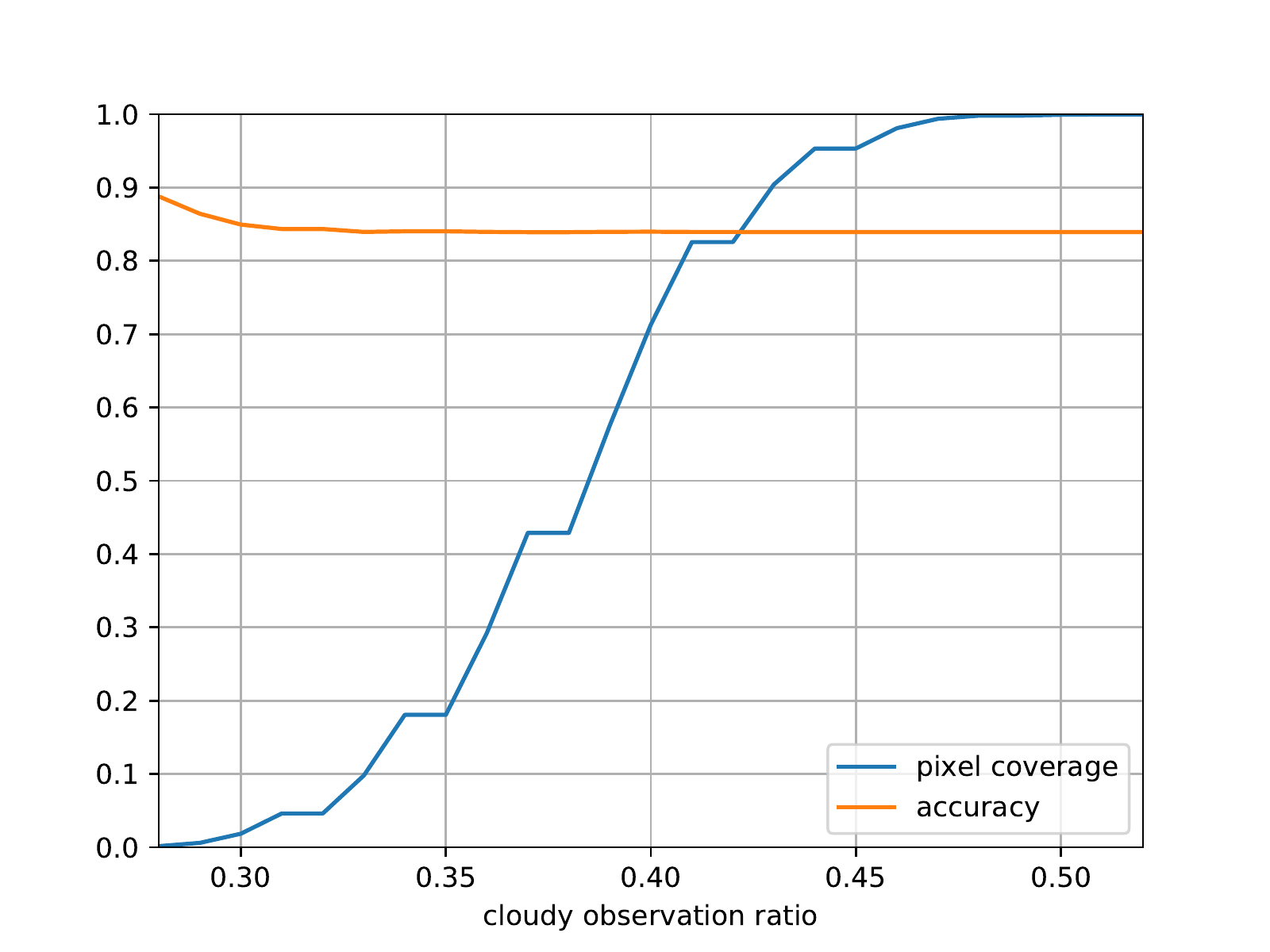}
        \caption{$c=0.5$}
    \end{subfigure}  
    \caption{Robustness against clouds. We plot classification accuracy against the proportion of cloud-free observations. Along the $x$-axis, pixels are ordered according to how often they are obscured by clouds. The blue curve depicts the cumulative number of pixels with at most a certain fraction of cloudy views. The orange curve shows the accuracy of the model evaluated over those pixels. The test was repeated with two different thresholds $c$ for the Sen2Cor cloud scores.}\label{fig:cloud_robustness}
\end{figure}

RNNs normally exhibit good robustness against missing data due to clouds, in the context of crop mapping see, e.g.,
\cite{marc_cloud}. 
%
We have conducted two experiments to further analyse robustness against clouds. For Table \ref{table:cloud_masking} we have trained the proposed model with the cloud mask as an additional input channel, so as to explicitly give the model the information which pixels are obscured by clouds. The performance is practically the same, suggesting that the model learns to detect the presence of clouds, such that the additional input brings no benefit.
In Figure~\ref{fig:cloud_robustness} we look at the issue from another angle: we thresholding the Sen2Cor cloud score to determine how many images are cloudy in the time series of every pixel. Then, we quantify how the varying number of cloud-free observations influences the performance. To that end we rank the classified pixels according to the number of cloudy observations in the time series. As can be seen, the accuracy remains almost constant as pixels are affected increasingly by cloud cover.

\begin{table}[th]
\begin{center}
\begin{tabular}{l|ccc|c}
\toprule
{\bf Method} &{\bf Prec (\%)} &{\bf Rec (\%)} &{\bf F1 (\%)} & {\bf Acc (\%)}\\
\midrule

{ms-convSTAR}  &{59.8} &{\textbf{49.7}} &{52.1}&{\textbf{88.0}}\\
{ms-convSTAR w/ CM}  &{\textbf{60.2}} &{48.8} &{\textbf{52.2}} &{87.9}
\\ \bottomrule 
\end{tabular}
\end{center}
\caption{Performance comparison: the proposed model vs.\ the same model with the cloud masks (CM) as additional input. Numbers are averaged over 5 cross-validation folds. The best score for each metric is shown in \textbf{bold}.}
\label{table:cloud_masking}
\end{table}

%% file: 06_discussion.tex

We go on to discuss limitations of the proposed approach, and  highlight cases where prediction is particularly difficult in the \emph{ZueriCrop} dataset.

Although the hierarchical approach significantly reduces the misclassification of underrepresented classes, exceedingly rare classes (e.g., \emph{beets}, \emph{lupine}, \emph{grain}) still often get confused with more frequent ones.
The full confusion matrix is shown in Fig.~\ref{fig:cm}. In most cases, very rare crops are misclassified as \emph{meadow}, which is the largest class in the dataset, with almost half of the labeled pixels.
The misclassification of field crops like \emph{grains} and \emph{lupine} may in part be attributed to a strong presence of weeds, which is especially common in organic or extensive cropping systems \citep{Giller2009, Gomiero2011}. This is also reflected by the \emph{mixed crop} class that represents a mixture of grain and legume crops used for Nitrogen efficient production of animal forage. Another source of errors are mixed pixels that span several crop types. Because the Sentinel-2 ground sampling distance of 10 meters is still relatively large relative to the small field sizes of Switzerland (see Fig.~\ref{fig:data_region}), mixed pixels can have a noticeable impact at field boundaries, where stripes of meadow (grassland) and hedges are common (see Fig.~\ref{fig:9_vs_4_ch}). 
Other examples of misclassifications are apple and pear, which belong to the same genus (family of Rosaceae). Similarly, sugar beets and beets (forage or vegetable) are the same species and just differentiated in the cultivar or variety.
Finally, 
intermediate crops, cover crops and secondary crops are not part of the reference dataset, but are of course visible in the satellite imagery throughout the year. It is required by Swiss law to grow cover crops during winter on bare soils as part of sustainable practice to reduce environmental degradation, such as erosion caused by strong rainfall events or nutrient leaching during winter~\citep{Prasuhn2012}. These intermediate or secondary crops are often, but not always, one of the sub-classes of the \emph{grassland} class. We hypothesize that this might be another reason for very rare classes being wrongly associated with \emph{grassland}.

Of course our proposed approach also has  limitations, which we discuss in the following, together with ideas how to mitigate them in future work. 
%
Like many deep neural networks, the model requires a significant amount of training labels for optimal performance. 
One possible way to alleviate the consequences of label scarcity during training is to use an active learning scheme. I.e., iterate between model training and the selection of maximally informative samples to extend the training set (which must then be annotated by an expert). Active learning can significantly reduce the amount of training labels needed to reach a defined performance level, the price to pay is that the overall lower annotation effort becomes more protracted and less projectable. 
%
Also, data-driven models are intrinsically tied to the data used during training. A domain shift, caused for instance by a change of target crop types, will require a re-training or fine-tuning of the model. At least fine-tuning may also be required to maintain performance if the model is applied in an area with different ecological background conditions. Pairing our method with a meta-learning scheme, as recently explored in~\citep{russwurm2020meta}, could ease the adaptation effort to different crop types and different geographic regions. 
%
Finally, the label hierarchy in this work is specific to Switzerland and different hierarchies will be needed for other agricultural regimes. In that context we reiterate that hard-coding a specific label hierarchy into the network architecture is an integral part of the current design, but lacks flexibility when moving to other environments. A technically challenging, but potentially useful extension to the current method could be to learn the label hierarchy from data rather than impose it a priori, using for instance unsupervised learning techniques. It can be expected that constructing the hierarchy in a data-driven manner, rather than with expert knowledge, will require fairly large amounts of data.




%% file: 07_conclusion.tex
We have proposed a novel, hierarchical classification approach for multi-temporal crop classification from satellite images -- in our case Sentinel-2, but the scheme is generic and could be applied to other sensors. Our ms-convSTAR method is a multi-stage convolutional recurrent neural network that leverages an explicit, hierarchical tree structure of the labels. Besides classifying the input data simultaneously at multiple interdependent hierarchy levels of granularity, the method also features an CNN-based label-refinement component to favour consistency across the hierarchy.

In our study, the labels are based on the classes of the Swiss governmental reporting scheme, and the hierarchy was defined by domain experts according to agronomic knowledge and best practices. 
Based on that hierarchy we also have collected a new dataset of Sentinel-2 time series images, \emph{ZueriCrop}, that densely covers a large agricultural region in Central Europe. The dataset is larger, more imbalanced, and more representative of real applications than earlier science datasets.
In the experiments on \emph{ZueriCrop}, ms-convSTAR has shown improved per-class performance and outperforms competing state-of-the-art methods. In particular, it greatly improves the classification of many rare classes.



%% file: 08_appendix.tex
\section{Performance comparison: Without polygon aggregation }
We provide models' performances  when polygon aggregation, i.e., majority voting is not performed as post-processing, in Table~\ref{table:no_majority_voting}. Compared to results in Table~\ref{table:result_sota}, scores for each method are decreased in a similar proportion especially in terms of overall accuracy and F1-score.

\begin{table}[th]
\footnotesize
\begin{center}
\tabcolsep=0.05cm
\scalebox{1.2}{
\begin{tabular}{l|ccc|c}
\toprule
{\bf Method}  &{\bf Prec (\%)} &{\bf Rec (\%)} &{\bf F1 (\%)}&{\bf Acc (\%)}  \\
\midrule
{Random Forest*}     &{43.6} &{\underline{36.2}} &{32.6} &{69.8}  \\
{LSTM \citep{marc_lstm}}     &{33.9} &{25.2} &{26.5} &{80.2}  \\
{TCN \citep{tempCONV}}     &{42.1} &{25.4} &{27.2} &{80.5}  \\
{Transformer \citep{breizhcrops}}     &{\textbf{53.7}} &{34.7} &{\underline{38.7}} &{82.0}  \\
{2D-CNN (U-Net)}     &{33.8} &{23.8} &{24.6} &{79.6}  \\
{U-Net+convLSTM \citep{fcn}}     &{44.1} &{31.2} &{31.7} &{81.1} \\
{Bi-convGRU \citep{bigru}}      &{49.1} &{35.7} &{38.4} &{\underline{82.4}} \\
\midrule
{ms-convSTAR}  &{\underline{53.0}} &{\textbf{44.2}} &{\textbf{46.2}}&{\textbf{83.9}} 
\\ \bottomrule 
\end{tabular} }
\end{center}
\caption{\textbf{Without polygon aggregation:} Performance comparison of ms-convSTAR (bottom row) with state-of-the-art methods when polygon aggregation, i.e., majority voting, is not performed as post-processing. Precision, recall and F1-score are mean values over all classes. All numbers are averaged over 5 cross-validation folds. The best score for each metric is shown with \textbf{bold} and the second best is \underline{underlined}. *Random Forest is trained with a balanced dataset by under-sampling the majority classes which leads to improved class-wise performance.} 
\label{table:no_majority_voting}
\end{table}

\section{Number of Input Channels}
\label{sec:num_input_ch}
In order to motivate our decision to use exclusively the 10m bands we conducted a simple experiment where we compare the performance of our method and of a Random Forest (RF) when using exclusively the 4 10m bands, and when using both the 10m and 20m bands. Results are reported in Table~\ref{table:channel}. As it can be seen, at least for the considered dataset, the improvements are marginal for the RF, and we actually observe a small performance drop for our method when using all the 9 bands. %
In terms of visual results, prediction comparison between 4 and 9 bands model are shown in Fig.~\ref{fig:cnn}. As can be seen the main difference is that the 9 bands model have some pixels that are misclassified on the edges of the fields. It is difficult to judge whether in this case the label is not fully accurate on the edges of the fields, or if the prediction is not accurate because of the limited resolution of the additional bands.

\begin{table}[t]
\begin{center}
\begin{tabular}{l|ccc|c}
\toprule
{\bf Method} &{\bf Prec (\%)} &{\bf Rec (\%)} &{\bf F1 (\%)} & {\bf Acc (\%)}\\
\midrule
{RF (9 channels)}     &{46.5} &{40.9} &{39.0} &{78.9}  \\
{RF (4 channels)}     &{46.4} &{40.7} &{38.9} &{78.8}  \\
\midrule
{ms-convSTAR (9 channels)}  &{56.3} &{46.6} &{48.7} &{87.8}\\
{ms-convSTAR (4 channels)}  &{\textbf{59.8}} &{\textbf{49.7}} &{\textbf{52.1}}&{\textbf{88.0}}
\\ \bottomrule 
\end{tabular}
\end{center}
\caption{Performance comparison w.r.t.\ number of input channels. 5-fold evaluation (bi-cubic upsampling of all channels with 20 meter meter GSD).}
\label{table:channel}
\end{table}

\begin{figure}[th]
    \centering
        \includegraphics[width=1.\columnwidth]{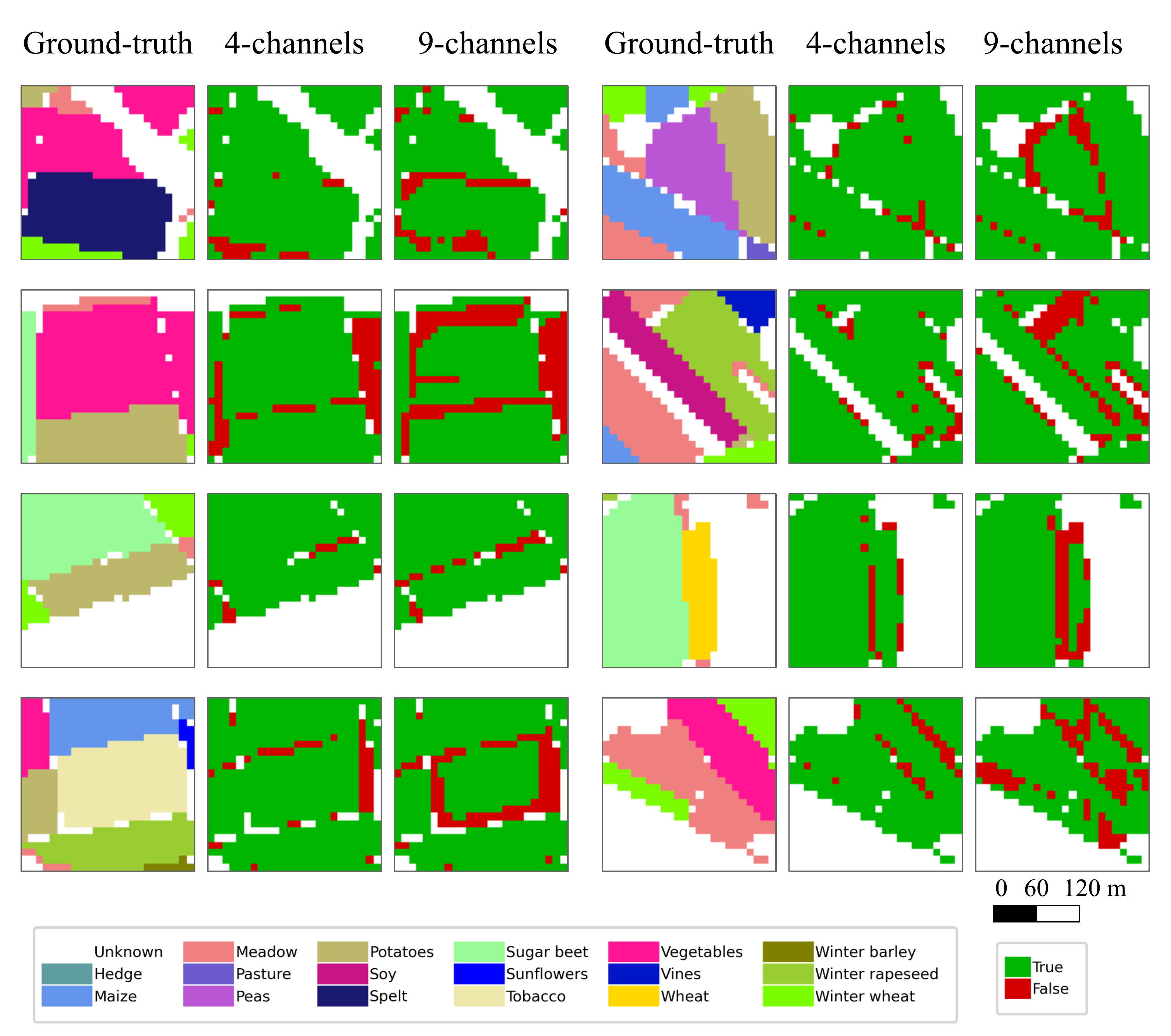}
    \caption{Qualitative comparison: Example failure cases of the 4-channels model vs the 9-channels model. Polygon aggregation is not performed as post-processing. Both models often fail at edge cases; however, the 9-channels model makes more mistakes at field edges.}
    \label{fig:9_vs_4_ch}
\end{figure}

%

%

\begin{figure}[th]
    \centering
        \includegraphics[width=1.\columnwidth]{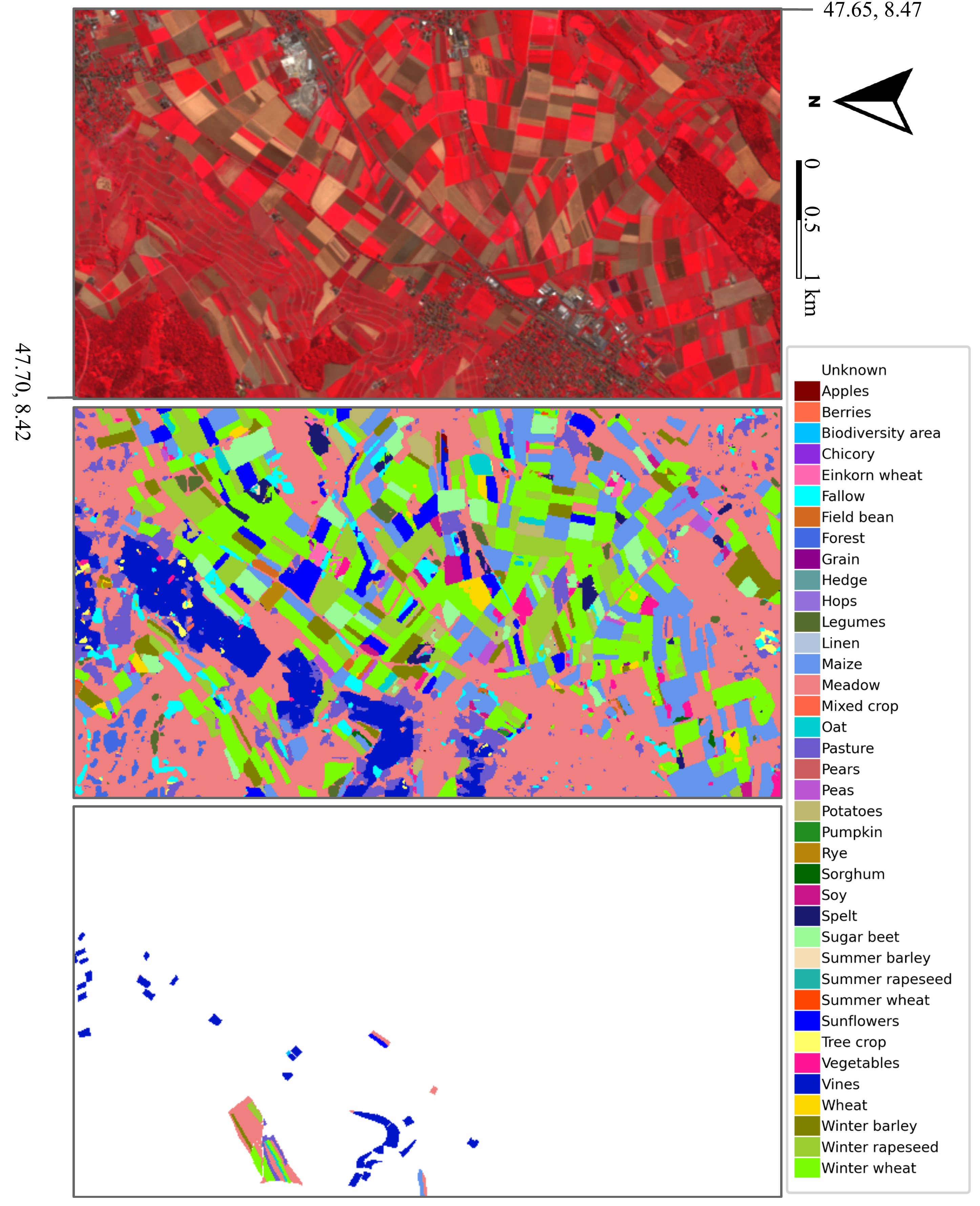}
    \caption{Qualitative result: The first row shows the Sentinel-2 satellite image (false color composite: NIR-Red-Green) of the test area, the second row shows the crop map generated by the proposed ms-convSTAR, and the third row shows the ground truth map.}
    \label{fig:mapping}
\end{figure}

\section{Class-wise Performance}
Class-wise accuracy for each granularity is given in Table~\ref{table:class_wise_scores}.

\begin{table}
\footnotesize
\begin{center}
\tabcolsep=0.05cm
\scalebox{1.0}{
\begin{tabular}{ |c|c|c|c|c|c|}
 \hline
\textbf{Level-1} & \textbf{Acc (\%)} & \textbf{Level-2 } & \textbf{Acc (\%)} & \textbf{Level-3} &  \textbf{Acc (\%)}  \\ 
    \hline
    \multirow{32}{*}{\makecell{Field Crops\\ ($41.39\%$)}}  &  \multirow{32}{*}{98.2} & \multirow{11}{*}{\makecell{Small Grain\\ Cereal ($16.89\%$)}} & \multirow{11}{*}{97.7} &  Summer Wheat ($0.13\%$) & 38.0 \\ \cline{5-6}
    &&&& Winter Wheat ($10.77\%$)  & 95.7\\\cline{5-6} 
    &&&& Wheat ($1.20\%$)  & 39.4\\\cline{5-6} 
    &&&&  Einkorn Wheat ($0.14\%$) & 49.8\\\cline{5-6} 
    &&&&  Summer Barley ($0.05\%$) & 26.1\\\cline{5-6} 
    &&&& Winter Barley ($3.49\%$)  & 96.4\\\cline{5-6} 
    &&&&  Grain ($0.05\%$) & 21.7\\\cline{5-6} 
    &&&&  Rye ($0.14\%$) & 68.1\\\cline{5-6} 
    &&&& Spelt ($0.70\%$)  & 84.0\\\cline{5-6} 
    &&&&  Oat ($0.21\%$) & 76.4\\\cline{5-6} 
    &&&& Buckwheat  ($0.02\%$)  & 7.0\\\cline{5-6} \cline{3-4}
    
    && \multirow{2}{*}{\makecell{Large Grain\\ Cereal ($10.37\%$)}} & \multirow{2}{*}{94.3} & Maize ($10.30\%$) & 94.7\\\cline{5-6} 
    &&&& Sorghum ($0.07\%$) & 27.0\\\cline{5-6} \cline{3-4}
    
    && \multirow{3}{*}{\makecell{Vegetable\\ Crop ($2.85\%$)}} & \multirow{3}{*}{83.4} & Vegetables ($2.76\%$) & 82.3\\\cline{5-6} 
    &&&& Pumpkin  ($0.03\%$) & 14.2\\\cline{5-6}
    &&&& Chicory ($0.07\%$) & 30.8\\ \cline{5-6} \cline{3-4}

    && \multirow{15}{*}{\makecell{Broad Leaf\\ Row Crop ($11.20\%$)}} & \multirow{15}{*}{95.1} & Sugar Beet ($4.26\%$) & 97.2\\\cline{5-6} 
    &&&& Beets ($0.01\%$)  & 0.0\\\cline{5-6}
    &&&& Potatoes ($1.65\%$)  & 86.6\\ \cline{5-6} 
    &&&& Sunflowers ($1.05\%$) & 92.8\\ \cline{5-6} 
    &&&& Linen ($0.04\%$) & 54.4\\ \cline{5-6}
    &&&& Hemp  ($0.02\%$) & 3.1\\ \cline{5-6} 
    &&&& Soy ($0.37\%$) & 87.4\\ \cline{5-6} 
    &&&&  Winter Rapeseed ($3.21\%$) & 97.6\\ \cline{5-6}
    &&&& Summer Rapeseed  ($0.02\%$)  & 0.0\\ \cline{5-6} 
    &&&& Field Bean ($0.15\%$) & 88.3\\ \cline{5-6} 
    &&&&  Peas ($0.32\%$) & 82.4\\ \cline{5-6} 
    &&&& Lupine ($<0.01\%$) & 0.0\\ \cline{5-6} 
    &&&& Tobacco  ($0.03\%$) & 48.5\\ \cline{5-6} 
    &&&& Mustard ($<0.01\%$) & 0.0\\ \cline{5-6} 
    &&&& Legumes ($0.05\%$) & 78.0\\ \cline{5-6} \cline{3-4}
    && \multirow{1}{*}{Crop Mix ($0.08\%$)} & \multirow{1}{*}{16.7} & Crop Mix ($0.08\%$) & 39.7\\
    
    \hline

    \multirow{3}{*}{\makecell{Grassland\\ ($56.00\%$)}}  &  \multirow{3}{*}{97.9} & Meadow ($47.76\%$) & 95.3 &  Meadow ($47.76\%$) & 94.9 \\ \cline{5-6}  \cline{3-4}
    && Pasture ($8.20\%$) & 37.4  &Pasture ($8.20\%$)  & 42.4\\\cline{5-6}  \cline{3-4}
    && Biodiversity A. ($0.04\%$) & 0.0  & Biodiversity A. ($0.04\%$)  & 0.0\\
    \hline

    \multirow{7}{*}{\makecell{Orchards\\ ($1.54\%$)}} & \multirow{7}{*}{49.6} & \multirow{6}{*}{Orchard Crop ($1.41\%$)} & \multirow{6}{*}{76.3} &  Apples ($0.40\%$) & 69.3 \\ \cline{5-6} 
    &&& &Pears ($0.07\%$) & 2.3\\\cline{5-6} 
    &&& &Vines ($0.83\%$)  & 82.2\\\cline{5-6} 
    &&& &Hops ($0.01\%$) & 0.0\\\cline{5-6} 
    &&& &Stone Fruit ($0.09\%$) & 24.1\\\cline{5-6} 
    &&& &Chestnut ($<0.01\%$)  & 0.0\\ \cline{5-6} \cline{3-4}
    && \multirow{1}{*}{Tree Crop ($0.13\%$)} & \multirow{1}{*}{20.9} & Tree Crop ($0.13\%$) & 37.2\\
    \hline
   
    \multirow{5}{*}{\makecell{Special Crops\\ ($1.00\%$)}}  &  \multirow{5}{*}{8.3} & \multirow{3}{*}{\makecell{Hedge, Gardens,\\  Multiple ($0.43\%$)}} & \multirow{3}{*}{1.0} &  Hedge ($0.42\%$) & 1.5\\\cline{5-6}
    &&&  & Gardens ($<0.01\%$) & 0.0\\\cline{5-6}
    &&& & Multiple  ($0.01\%$) & 0.0\\\cline{5-6} \cline{3-4}
    && \multirow{1}{*}{Berries ($0.22\%$)} & \multirow{1}{*}{18.8} & Berries ($0.22\%$) & 25.8\\\cline{5-6} \cline{3-4}
    && \multirow{1}{*}{Fallow ($0.36\%$)} & \multirow{1}{*}{65.8} & Fallow ($0.36\%$) & 70.3\\
    
    \hline 
    
    \multirow{1}{*}{Forest ($0.07\%$)} & \multirow{1}{*}{0.0} & \multirow{1}{*}{Forest ($0.07\%$)} & \multirow{1}{*}{0.0} &  Forest ($0.07\%$) & 1.6 \\
    \hline
    
\end{tabular}
}
\end{center}
\caption{Class-wise performance of proposed ms-convSTAR at different levels of granularity. All numbers are averaged over 5 cross-validation folds. Class frequencies are given in parenthesis.} 
\label{table:class_wise_scores}
\end{table}

\section{More Details about Baselines}
Parameters of baseline models and links for their source codes are given in Table~\ref{table:params} and Table~\ref{table:source}, respectively.

\begin{table}[h]
\begin{center}
\begin{tabular}{ |c|c|c|c|c|m{3.2cm}|} 
 \hline
 \textbf{Method} & \textbf{Hidden} & \textbf{Layer} & \textbf{Kernel} & \textbf{Batch} &  \textbf{Note}  \\ 
 \hline
 Random Forest &  &  & &  &  {max\_depth:60 max\_features:auto min\_sample\_leaf:1 min\_sample\_split:3 n\_estimators:1000}\\ 
 \hline
 LSTM & 128 & 1 & & 576 & \\  
 \hline
 TCN & 64 & 3 & 5 & 576 & dropout: 0.5 \\  
  \hline
 Transformer & 64 & 4 &  & 576 & head: 4 \\ 
   \hline
 Bi-convGRU & 64 & 1 & 3x3 & 4 & \\ 
    \hline
 U-Net &  & 12 & 3x3 & 4 & filters: 64, 64, 128, 128, 256, 256 \\ 
     \hline
 U-Net+convLSTM & 256 & 12+1 & 3x3 & 4 &filters: 64, 64, 128, 128, 256, 256\\ 
 \hline
\end{tabular}
\end{center}
\caption{Parameters of baseline methods.} 
\label{table:params}
\end{table}

\begin{table}[h]
\begin{center}
\tabcolsep=0.05cm
\begin{tabular}{ |c|c|} 
 \hline
 \textbf{Method} &  \textbf{Source} \\ 
 \hline
 Random Forest &  \url{https://scikit-learn.org}\\ 
\hline
 LSTM &  \url{https://pytorch.org/docs/stable/nn.html}\\ 
 \hline
 TCN &  \url{https://github.com/charlotte-pel/temporalCNN}\\ 
 \hline
 Transformer &  \url{https://github.com/dl4sits/BreizhCrops}\\
 \hline
 Bi-convGRU &  \url{https://github.com/TUM-LMF/MTLCC}\\
\hline
 U-Net &   \url{https://github.com/roserustowicz/crop-type-mapping}\\
 \hline
 U-Net+convLSTM &  \url{https://github.com/roserustowicz/crop-type-mapping}\\
 \hline
\end{tabular}
\end{center}
\caption{Code sources of baseline methods. } 
\label{table:source}
\end{table}